\documentclass[10pt,twocolumn,letterpaper]{article}

\usepackage{cvpr}              

\usepackage{graphicx}
\usepackage{amsmath}
\usepackage{amssymb}
\usepackage{booktabs}
\usepackage{physics}
\usepackage{tabularx}
\usepackage{textcomp}
\usepackage[T1]{fontenc}
\usepackage[accsupp]{axessibility} 
\usepackage[pagebackref,breaklinks,colorlinks]{hyperref}
\usepackage[table]{xcolor}

\usepackage{bbding}
\usepackage[capitalize]{cleveref}
\usepackage[table]{xcolor}
\crefname{section}{Sec.}{Secs.}
\Crefname{section}{Section}{Sections}
\Crefname{table}{Table}{Tables}
\crefname{table}{Tab.}{Tabs.}
\newcommand{\Tref}[1]{Table~\ref{#1}}

\newcommand{\Fref}[1]{Fig.~\ref{#1}}

\def\cvprPaperID{498} 
\def\confName{CVPR}
\def\confYear{2023}

\begin{document}

\title{Scalable, Detailed and Mask-Free Universal Photometric Stereo}
\author{Satoshi Ikehata\\
National Institute of Informatics (NII)\\
{\tt\small sikehata@nii.ac.jp}
}
\twocolumn[{%
\renewcommand\twocolumn[1][]{#1}%
\maketitle
\vspace{-30pt}
\begin{center}
    \centering
    \captionsetup{type=figure}
    \includegraphics[width=160mm]{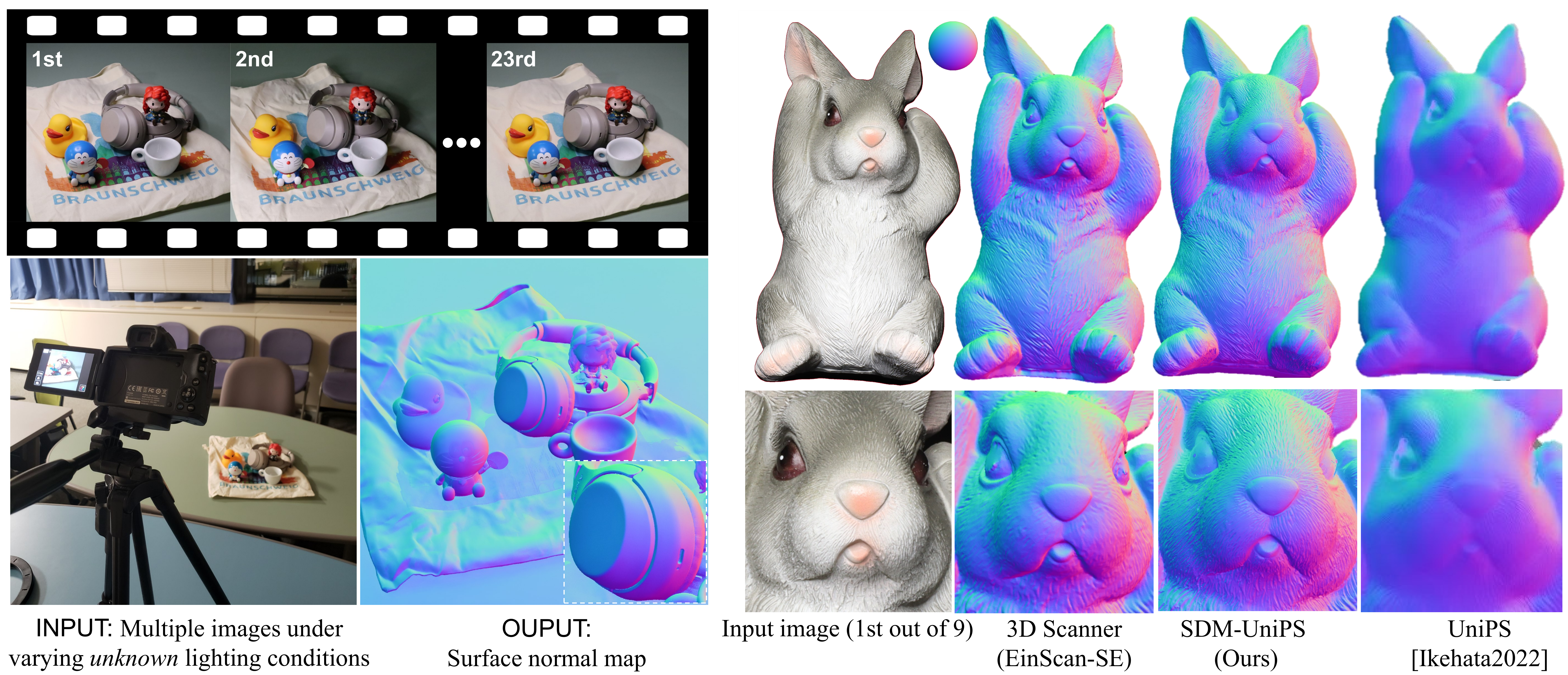}
    \vspace{-10pt}
    \captionof{figure}{Given multiple images under unknown spatially-varying illuminations, our method can recover the detailed surface normal map of non-convex, non-Lambertian surfaces~(Left). Our method even surpasses the level of detail provided by consumer 3-D scanners~(Right).}
    \label{fig:teaser}
\end{center}%
}]
\begin{abstract}
In this paper, we introduce SDM-UniPS, a groundbreaking Scalable, Detailed, Mask-free, and Universal Photometric Stereo network. Our approach can recover astonishingly intricate surface normal maps, rivaling the quality of 3D scanners, even when images are captured under unknown, spatially-varying lighting conditions in uncontrolled environments. We have extended previous universal photometric stereo networks to extract spatial-light features, utilizing all available information in high-resolution input images and accounting for non-local interactions among surface points. Moreover, we present a new synthetic training dataset that encompasses a diverse range of shapes, materials, and illumination scenarios found in real-world scenes. Through extensive evaluation, we demonstrate that our method not only surpasses calibrated, lighting-specific techniques on public benchmarks, but also excels with a significantly smaller number of input images even without object masks.
\end{abstract}
\vspace{-20pt}
\section{Introduction}\label{sec:intro}
Photometric stereo~\cite{Woodham1980} aims to deduce the surface normal map of a scene by analyzing images captured from a fixed perspective under diverse lighting conditions. Until very recently, all photometric stereo methods assumed their specific lighting conditions, which led to limitations in their applicability. For instance, methods that assumed directional lighting conditions (\eg, \cite{Ikehata2012,Ikehata2014a,Ikehata2018}) were unsuitable under natural illumination, and vice versa (\eg,~\cite{Mo2018,Haefner2019}).

To overcome this limitation, the ``universal'' photometric stereo method (UniPS)~\cite{Ikehata2022} has been introduced, designed to operate under unknown and arbitrary lighting conditions. In contrast to prior uncalibrated photometric stereo methods~\cite{Chen2019,Chen2020,Kaya2021}, which assumed specific physically-based lighting models, this method encodes a non-physical feature at each pixel for representing spatially-varying illumination, which is served as a substitute for physical lighting parameters within the calibrated photometric stereo network~\cite{Ikehata2021}. This method has taken the first step towards dealing with unknown, spatially-varying illumination that none of the existing methods could handle.
\renewcommand{\thefootnote}{\fnsymbol{footnote}}
\footnote[0]{Supported by JSPS KAKENHI Grant Number 22K17919.}
\renewcommand{\thefootnote}{\arabic{footnote}}
However, the surface normal map recovered by UniPS, while not entirely inaccurate, appears blurry and lacks fine detail (see the top-right corner of \Fref{fig:teaser}). Upon investigation, we pinpointed three fundamental factors contributing to the subpar reconstruction performance. Firstly, extracting illumination features (\ie, global lighting contexts) from downsampled images caused a loss of information at higher input resolutions and produced blurry artifacts. Secondly, UniPS employs a pixel-wise calibrated photometric stereo network to predict surface normals using illumination features, which leads to imprecise overall shape recovery. Although pixel-wise methods~\cite{Ikehata2021,Ikehata2014a,Ikehata2018} offer advantages in capturing finer details compared to image-wise methods~\cite{Chen2018,Taniai2018,Li2022a}, they suffer from an inability to incorporate global information.

Lastly, the third issue lies in the limited variety of shape, material, and illumination conditions present in the training data, which hampers its capacity to adapt to a diverse range of real-world situations. This limitation primarily stems from the fact that current datasets (\ie, PS-Wild~\cite{Ikehata2022}) do not include renderings under light sources with high-frequency components focused on specific incident angles, such as point or directional sources. Consequently, the method exhibits considerable performance degradation when exposed to directional lighting setups like DiLiGenT~\cite{Shi2016}, as will be demonstrated later in this paper.

In this paper, we present a groundbreaking photometric stereo network, the Scalable, Detailed, and Mask-Free Universal Photometric Stereo Network (SDM-UniPS), which recovers normal maps with remarkable accuracy from images captured under extremely uncontrolled lighting conditions. As shown in~\Fref{fig:teaser}, SDM-UniPS is {\it scalable}, enabling the generation of normal maps from images with substantially higher resolution (\eg, 2048x2048) than the training data (\eg, 512x512); it is {\it detailed}, providing more accurate normal maps on DiLiGenT~\cite{Shi2016} with a limited number of input images than most existing orthographic photometric stereo techniques, including calibrated methods, and in some cases, surpassing 3D scanners in detail; and it is {\it mask-free}, allowing for application even when masks are absent, unlike many conventional methods.
Our technical novelties include:
\begin{enumerate}
\item The development of a \textit{scale-invariant spatial-light feature encoder} that efficiently extracts illumination features while utilizing all input data and maintaining scalability with respect to input image size. Our encoder, based on the "split-and-merge" strategy, accommodates varying input image sizes during training and testing without sacrificing performance.

\item The development of a surface normal decoder utilizing our novel \textit{pixel-sampling transformer}. By randomly sampling pixels of fixed size, we simultaneously predict surface normals through non-local interactions among sampled pixels using Transformers~\cite{Vaswani2017}, effectively accounting for global information.

\item The creation of a new synthetic training dataset, comprising multiple objects with diverse textures within a scene, rendered under significantly varied lighting conditions that include both low and high-frequency illuminations.
\end{enumerate}
We believe that the most significant contribution is the extraordinary time savings from data acquisition to normal map recovery compared to existing photometric stereo algorithms requiring meticulous lighting control, even in the uncalibrated setup. This progress allows photometric stereo to be executed at home, literally ``in the wild'' setup.
\section{Related Works}
In this section, we provide a succinct overview of photometric stereo literature focusing on the single orthographic camera assumption. Alternative setups (\eg, perspective, multi-view cameras) are beyond the scope of this work.
\vspace{4pt}\\
\noindent \textbf{Optimization-based Approach:}
The majority of photometric stereo methods assume calibrated, directional lighting following Woodham~\cite{Woodham1980} and optimize parameters by inversely solving a physics-based image formation model. This approach can be further categorized into robust methods, where non-Lambertian components are treated as outliers~\cite{Mukaigawa2007, Yu2010, Wu2010,Ikehata2012}; model-based methods, which explicitly account for non-Lambertian reflectance~\cite{Goldman2005,Shi2012a,Ikehata2014b}; and example-based methods~\cite{Silver1980,Hertzmann2005,Hui2017} that leverage the observations of known objects captured under identical conditions as the target scene. The uncalibrated task is akin to the calibrated one, but with unknown lighting parameters. Until recently, most uncalibrated photometric stereo algorithms assumed Lambertian integrable surfaces and aimed to resolve the General Bas-Relief ambiguity~\cite{Hayakawa1994,Drobohlav2002,Alldrin2007a,Shi2010,Favaro2012,Wu2013,Papadhimitri2013,Feng2013}. In contrast to these works, photometric stereo under natural lights has also been explored, wherein natural illumination is approximated using spherical harmonics~\cite{Basri2007,Haefner2019}, dominant sun lighting~\cite{Ackermann2012,Hold2019}, or equivalent directional lighting~\cite{Mo2018,Guo2021}. Although most optimization-based methods do not require external training data, they are fundamentally limited in handling global illumination phenomena (\eg, inter-reflections) that cannot be described by the predefined point-wise image formation model.
\vspace{4pt}\\
\noindent \textbf{Learning-based Approach:}
Learning-based methods are effective in addressing complex phenomena that are challenging to represent within simple image formation models. However, the first photometric stereo network~\cite{Santo2017} necessitated consistent lighting conditions during both training and testing. To address this limitation, various strategies have been investigated, such as observation maps~\cite{Ikehata2018, Logothetis2021}, set-pooling~\cite{Chen2018, Yakun2021}, graph-convolution~\cite{Yao2020}, and self-attention~\cite{Liu2021, Ikehata2021}. Furthermore, researchers have explored uncalibrated deep photometric stereo networks~\cite{Chen2019, Chen2020, Kaya2021, Tiwari2022}, where lighting parameters and surface normals are recovered sequentially. Self-supervised neural inverse rendering methods have been developed without the need for external data supervision. Taniai and Maehara~\cite{Taniai2018} used neural networks instead of parametric physical models, with images and lighting as input. This work was expanded by Li and Li~\cite{Li2022a, Li2022b}, who incorporated recent neural coordinate-based representations~\cite{Mildenhall2020}. However, despite their tremendous efforts, these methods are designed to work with only single directional light source and have limited ability to generalize to more complex lighting environments.
\vspace{4pt}\\
\noindent \textbf{Universal Photometric Stereo Network:}
The universal photometric stereo network (UniPS)~\cite{Ikehata2022} was the first to eliminate the prior lighting model assumption by leveraging a non-physical lighting representation called global lighting contexts. These global lighting contexts are recovered for each lighting condition through pixel-wise communication of hierarchical feature maps along the light-axis using Transformers~\cite{Vaswani2017}. During surface normal prediction, a single location is individually selected, and the network aggregates all the global lighting contexts (bilinearly interpolated from the canonical resolution) and raw observations at the location under different lighting conditions to pixel-wise predict the surface normal. This method introduced two strategies to handle high-resolution images: down-sampling images to the canonical resolution for recovering global lighting contexts, and employing pixel-wise surface normal prediction. Although these two concepts contributed to the scalability of image size, they resulted in performance degradation due to the loss of input information and the absence of a non-local perspective, as previously discussed.

Our work draws inspiration from~\cite{Ikehata2022} and shares some fundamental ideas, particularly the use of Transformers~\cite{Vaswani2017} for communicating and aggregating features along the light-axis. However, our method diverges from~\cite{Ikehata2022} by fully utilizing input information in a non-local manner, which leads to a significant enhancement in reconstruction quality.
\section{Method}\label{sec:method}
We target the challenging universal photometric stereo task, which was recently introduced in~\cite{Ikehata2022}. Unlike prior calibrated and uncalibrated tasks, the universal task makes no assumptions about surface geometry, material properties, or, most importantly, lighting conditions. The objective of this task is to recover a normal map $N\in \mathbb{R}^{H\times W \times 3}$ from images $I_k\in \mathbb{R}^{H\times W\times 3}; k\in{1,\dots,K}$ captured under $K$ unknown lighting conditions using an orthographic camera. Optionally, an object mask $M\in \mathbb{R}^{H\times W}$ may be provided.

\begin{figure*}[!t]
	\begin{center}
		\includegraphics[width=160mm]{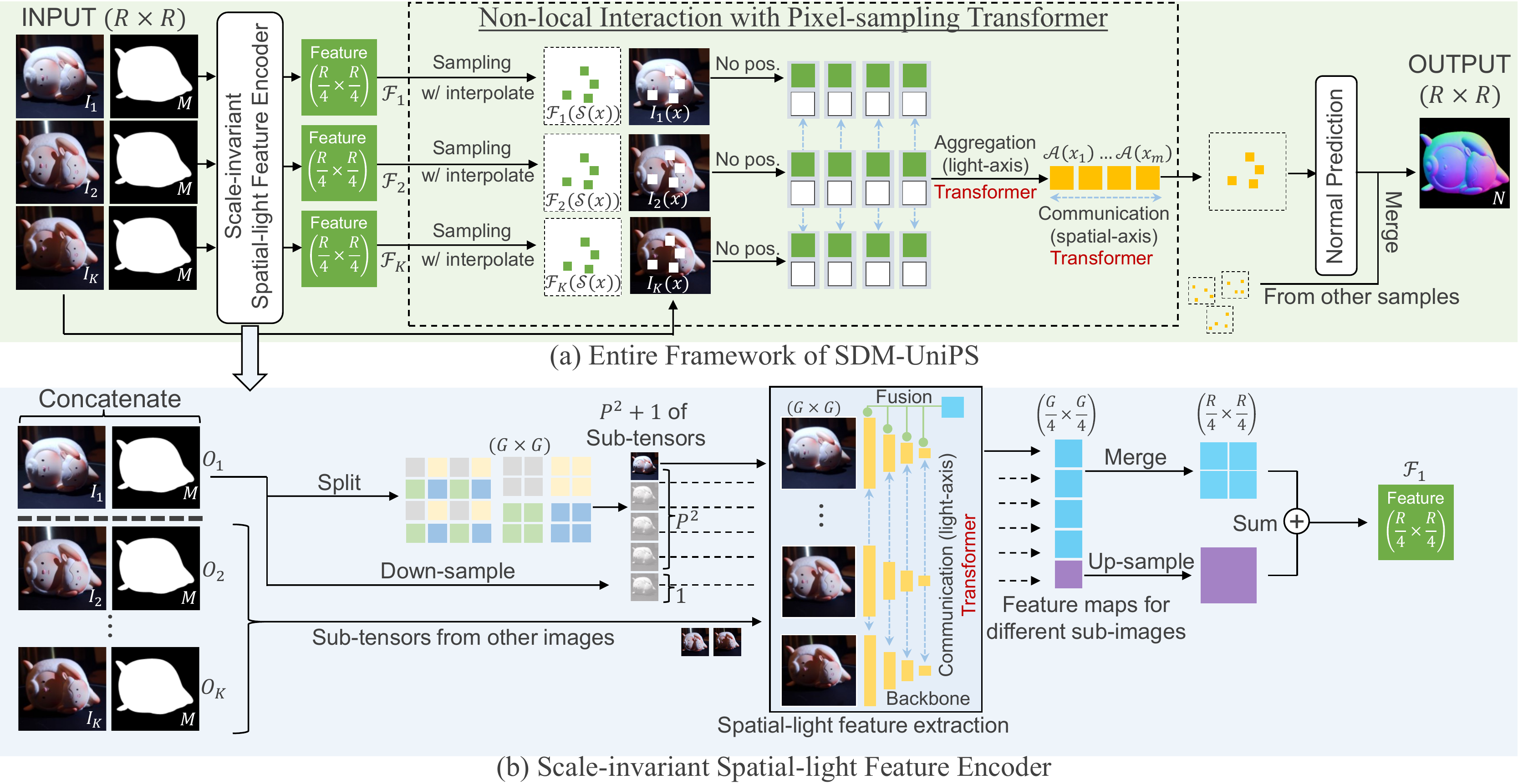}
	\end{center}
	\vspace{-15pt}
	\caption{Our entire framework is illustrated in (a). Given multiple images and an object mask (optional), the scale-invariant spatial-light encoder (detailed in (b)) extracts a feature map for each image. The surface normal vectors are independently recovered at each of pixel samples (\ie, 2048) after the non-local interaction among aggregated features from interpolated feature maps and raw observations.}
	\label{fig:architecture}
	\vspace{-5pt}
\end{figure*}

Our method (SDM-UniPS) is illustrated in~\Fref{fig:architecture}. Given pre-processed images and an optional object mask, feature maps for each lighting condition are extracted through interactions along the spatial and light axes (\ie, the scale-invariant spatial-light feature encoder). We then randomly sample locations from the coordinate system of the input image and bilinearly interpolate features at these locations. Features and raw observations at each location are aggregated pixel-wise, and surface normals are recovered from the aggregated features after non-local spatial interaction among them (\ie, the pixel-sampling Transformer). In line with~\cite{Ikehata2022}, we focus on describing high-level concepts rather than providing detailed explanations for the sake of clarity. Refer to the appendix for a comprehensive description of the network architectures.
\subsection{SDM-UniPS}
\noindent\textbf{Pre-processing:}
As in~\cite{Ikehata2022}, we resize or crop input images to a resolution ($R$) that is divisible by $32$, which is accepted by most hierarchical vision backbones. To ensure that image values are within a similar range, each image is normalized by a random value between its maximum and mean.
\vspace{4pt}\\
\noindent\textbf{Scale-invariant Spatial-light Feature Encoder:} After the pre-processing, we extract feature maps from images and an optional object mask through the interaction along both spatial and light axes. Following the basic framework in~\cite{Ikehata2022}, each image and object mask~\footnote{Without a mask, a matrix with all values set to one is concatenated.} are concatenated to form a tensor $O_k\in \mathbb{R}^{R\times R\times 4}$, which is then input to the common vision backbone~\cite{Liu2021Swin,Liu2022Swin,Liu2022} to extract hierarchical feature maps $B_k^s\in \mathbb{R}^{\frac{R}{S_s}\times \frac{R}{S_s}\times C_s},;s\in{1,2,3,4}$. Here, $S_s\in {4,8,16,32}$ represents the scale of the $s$-th feature map, and $C_s$ is the dimension of features at that scale. For each feature scale, features from different tensors at the same pixel interact with each other along the light-axis using naïve Transformers~\cite{Vaswani2017}. Finally, hierarchical feature maps are fused to $\mathcal{F}_k\in \mathbb{R}^{\frac{R}{4}\times \frac{R}{4}\times C_{\mathcal{F}}}$ using the feature pyramid network~\cite{Xiao2018}, where $C_{\mathcal{F}}$ is the output feature dimension. Note that, unlike~\cite{Ikehata2022}, we used a varying number of Transformer blocks at each hierarchy scale (\ie, the number of blocks changes from [1,1,1,1] to [0,1,2,4]) so that the deeper features interact more than the shallow ones.

In UniPS~\cite{Ikehata2022}, images and a mask are down-sampled to a \textit{canonical resolution} before being input to the backbone network. This resolution must be constant and sufficiently small (\eg, 256x256) to prevent excessive memory consumption during feature extraction, particularly when dealing with high-resolution input images. Additionally, using a constant resolution ensures that tensors of the same shape are fed to the backbone, which helps to avoid significant performance degradation due to a large discrepancy in input tensor shapes between training and testing. Consequently, down-sampling leads to the loss of much information in the input images, resulting in a blurry normal map recovery.

To address it, we propose a \textit{scale-invariant spatial-light feature encoder} designed to maintain a consistent, small input resolution for the backbone network while preserving information from input images. Specifically, instead of downsampling, we suggest \textit{splitting} the input tensor into non-overlapping sub-tensors with a constant, small resolution. In greater detail, we decompose $O$ into $P^2$ sub-tensors of size $G\times G$ ($G=256$ in our implementation, $P\triangleq R/G$) by taking a single sample from every $P\times P$ pixel and stacking them as sub-tensors, as illustrated in~\Fref{fig:architecture}. Each sub-tensor encompasses the entire portion of the original tensor but is slightly shifted. All sub-tensors are processed independently through the same spatial-light feature encoder and subsequently merged back into a tensor of size $(\frac{R}{4}\times \frac{R}{4}\times C_{\mathcal{F}})$. The combined feature maps from the sub-tensors retain all input information since no downsampling occurred. However, the absence of interaction among different sub-tensors leads to significant block artifacts, particularly when $P$ is large. To mitigate this, another feature map encoded from the naively downsized image is added\footnote{Concatenation is also possible, but it did not improve the results despite increased memory consumption.} to the merged feature maps, promoting interaction among sub-tensors. Optionally, when $P$ is larger than $4$, we apply depth-wise Gaussian filtering (\ie, kernel size is $P$-1) to the feature maps to further enhance the interaction. Finally, we obtain the scale-invariant spatial-light feature maps $\mathcal{F}_k\in \mathbb{R}^{\frac{R}{4}\times \frac{R}{4}\times C_{\mathcal{F}}}$ for every lighting condition.

\vspace{4pt}
\noindent\textbf{Non-local Interaction with Pixel-sampling Transformer:}
Given the scale-invariant spatial-light feature maps $\mathcal{F}_k$ and images $I_k$, the surface normal is recovered after pixel-wise feature aggregation along the light-axis (\ie, the light channel shrinks from $K$ to $1$). Feature aggregation under different lighting conditions is a fundamental step in photometric stereo networks, and various strategies have been studied, such as observation maps~\cite{Ikehata2018,Logothetis2021}, max-pooling~\cite{Chen2018,Chen2020}, graph-convolution~\cite{Yao2020}, and self-attention~\cite{Ikehata2021,Ikehata2022}. We utilize the Transformer model with self-attention\cite{Vaswani2017} as in the encoder following UniPS~\cite{Ikehata2022}. UniPS directly predicted surface normals from pixel-wise aggregated feature vectors, following other pixel-wise methods~\cite{Ikehata2021,Ikehata2014a,Ikehata2018}, without considering non-local interactions. However, aggregated features lose lighting-specific information, naturally obtaining lighting-invariant representations more related to surface attributes than those before aggregation. In traditional physics-based vision tasks, common constraints including isotropy~\cite{Alldrin2007a}, reciprocity symmetry~\cite{Tan2011}, reflectance monotonicity~\cite{Chandraker2011a}, sparse reflectance basis~\cite{Goldman2005}, and surface integrability~\cite{Onn1990} are mostly shared on the surface, not limited to a single surface point. Thus, considering non-local interactions of aggregated features at multiple surface points is crucial in physics-based tasks. 

Applying image-wise neural networks like CNNs on the aggregated feature map demands enormous computational cost for large output resolutions (\eg, $2048\times 2048$), and risks compromising output normal map details. To address these issues, we draw inspiration from recent Transformers on 3-D points~\cite{Zhao2021,Wu2022} and apply a Transformer on a fixed number ($m$) of {\it pixel samples} (\eg, $m=2048$) from random locations in the input coordinate system. We term this the {\it pixel-sampling Transformer}. Unlike image-based approaches, pixel-sampling Transformer's memory consumption is constant per sample set, scaling to arbitrary image sizes. Moreover, by applying the Transformer to a randomly sampled set of locations, local interactions that may lead to over-smoothing of feature maps (\eg, in CNNs) are almost entirely eliminated.

Concretely, given $m$ random pixels $x_{i=1,\dots,m}$ from the masked region of the input coordinate system, we interpolate features at those pixels as $\mathcal{F}_{{1,\dots,K}}(\mathcal{S}(x_i))$, where $\mathcal{S}$ is the bilinear interpolation operator. Then, interpolated features are concatenated with corresponding raw observations $I_{{1,\dots,K}}(x_i)$ and aggregated to $\mathcal{A}(x_i)$ with pooling by multi-head attention (PMA)\cite{Lee2019}, as in\cite{Ikehata2022}. Given aggregated features at different pixels in the same sample set, we apply another naïve Transformer~\cite{Vaswani2017} to perform non-local interactions. Since the goal of this process is to consider surface-level interactions based on physical attributes, pixel coordinate information is unnecessary. Thus, we {\it don't} apply position embeddings to samples, unlike most existing visual Transformer models (\textit{\eg}~\cite{Dosovitskiy2020,Liu2021Swin}), allowing the samples to propagate their aggregated features without location information.

After the non-local interaction, we apply a two-layer MLP to predict surface normals at sampled locations. Finally, surface normals for each set are merged to obtain the final surface normal map at the input image resolution. This pixel-sampling Transformer approach facilitates non-local interactions while maintaining computational efficiency and preserving output normal map details, making it suitable for physics-based tasks with high-resolution images.
\subsection{PS-Mix Dataset}\label{sec:dataset}
\begin{figure}[t]
	\begin{center}
		\includegraphics[width=85mm]{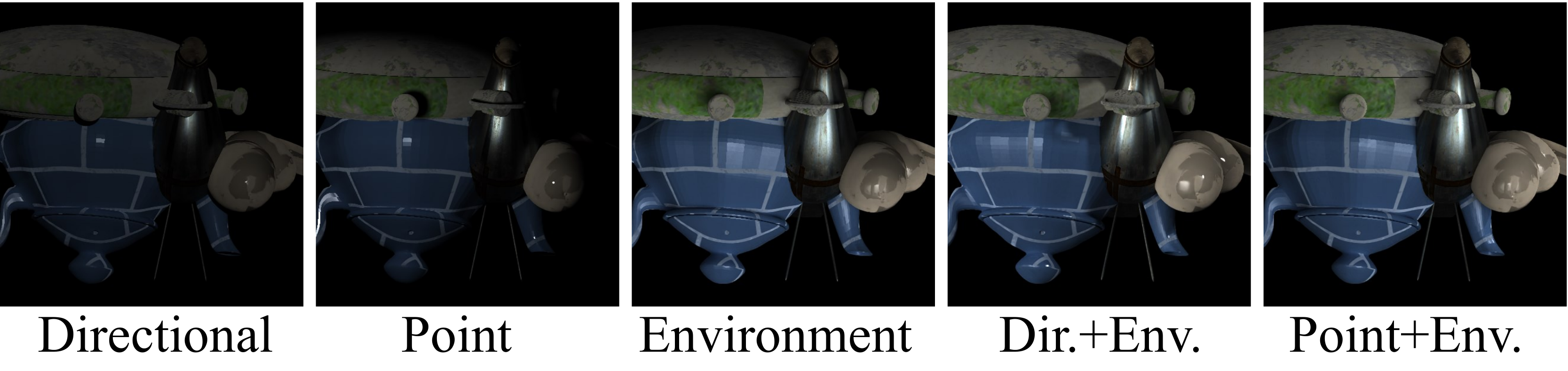}
	\end{center}
	\vspace{-20pt}
	\caption{Examples in PS-Mix under different lighting conditions.}
	\vspace{-10pt}
	\label{fig:dataset}
\end{figure}
To train their universal photometric stereo network, Ikehata~\cite{Ikehata2022} presented the PS-Wild training dataset, which rendered more than 10,000 scenes with commercial AdobeStock 3-D assets~\cite{AdobeStock}. One of the issues in PS-Wild is that each scene consists of only a single object of uniform material. Furthermore, the environment lighting used for rendering scenes in~\cite{Ikehata2022} rarely has high-frequency illumination (\eg, a single point light source); therefore, the rendered images are biased towards low-frequency lighting conditions.

In this paper, we create a new training dataset that solves the issues in the PS-Wild training dataset. Instead of putting a single object of uniform material in each scene, we put multiple objects that overlap with each other in the same scene and give them different materials. To ensure that the material category is diverse in a scene, we manually categorized 897 texture maps in the AdobeStock material assets into 421 diffuse, 219 specular, and 257 metallic textures. For each scene, we randomly select four objects from 410 AdobeStock 3-D models and assign three textures from all three material categories and randomly choose one for each object. Furthermore, to make the lighting conditions more diverse, instead of using only environment lighting to render images, we use five types of light source configurations and mix them to render one scene; (a) environment lighting, (b) single directional lighting, (c) single point lighting, (d) (a)+(b), and (e) (a)+(c). The direction and position of light sources are randomly assigned within the valid range of parameters\footnote{Light directions are selected from the upper unit hemisphere, and point light positions are selected inside the hemisphere.}. We followed PS-Wild~\cite{Ikehata2022} for other rendering techniques (\eg, auto-exposure, object scale adjustment). Our dataset consists of 34,921 scenes, and each scene is rendered to output 10 of 16-bit, 512$\times$512 images. In~\Fref{fig:dataset}, we show sample images under each lighting condition for the same scene.
\section{Results}

\begin{table}[!t]
\setlength{\tabcolsep}{1mm} 
    \centering
    \caption{Ablation study on PS-Wild-Test~\cite{Ikehata2022}.}
    \vspace{-10pt}
    \small
    \begin{tabularx}{75mm}{Xccccc}
    \toprule
        Method & Training & Dir. & HDRI & Dir.+HDRI\\
        \midrule
        I22 (UniPS)~\cite{Ikehata2022}&PS-Wild&17.0&14.5&13.8\\
        \midrule
        Only Local (baseline) & PS-Mix & 8.4&14.7&11.8\\
        +Non-local (32) & PS-Mix &7.8&14.9&10.8\\
        +Non-local (128) & PS-Mix &6.2&13.0&8.9\\
        +Non-local (512) & PS-Mix &5.8&12.4&8.2\\
        +Non-local (2048) & PS-Mix &5.7&12.2&8.0\\
        +Non-local (20480) & PS-Mix &5.7&12.3&8.0\\
        +Scale-invariant Enc.& PS-Mix&4.8&11.1&7.5\\
    \bottomrule
    \end{tabularx}
    \label{table:ablation}
\end{table}
\begin{figure}[t]
	\begin{center}
		\includegraphics[width=85mm]{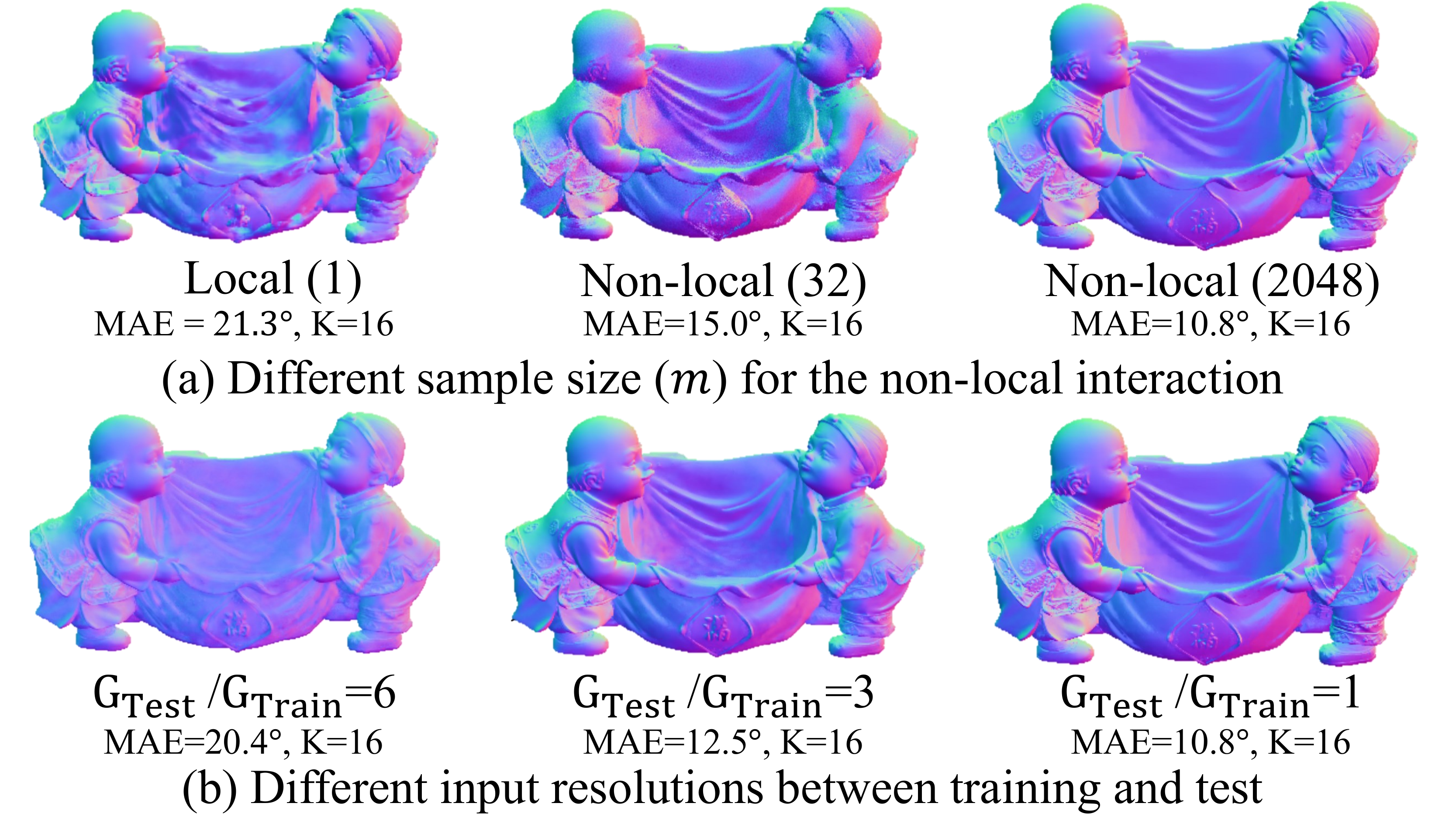}
	\end{center}
	\vspace{-10pt}
	\caption{(a) Comparison of different sample size ($m$) for the non-local interaction in the normal prediction. (b) Comparison of different input resolutions to the encoder between training and test.}
	\vspace{-10pt}
	\label{fig:ablation}
\end{figure}

\begin{table*}[!t]
\setlength{\tabcolsep}{1mm} 
    \centering
    \caption{Evaluation on DiLiGenT~\cite{Shi2016} (Mean Angular Errors in degrees). All 96 images were used except where K is shown.}
    \vspace{-10pt}
    \small
    \begin{tabularx}{165mm}{Xccccccccccccc}
    \toprule
         Method & Approach &Task& Ball & Bear &Buddha & Cat & Cow & Goblet & Harvest & Pot1 & Pot2 & Reading & Ave. \\
         \midrule
        Woodham~\cite{Woodham1980}&Self-Sup.&Calibrated&4.1&8.4&14.9&8.4&25.6&18.5&30.6&8.4&14.7&19.8&15.3\\
        IW14~\cite{Ikehata2014a}&Self-Sup.&Calibrated&2.0&4.8&8.4&5.4&13.3&8.7&18.9&6.9&10.2&12.0&9.1\\
        IA14~\cite{Ikehata2014b}&Self-Sup.&Calibrated&3.3&7.1&10.5&6.7&13.1&9.7&26.0&6.6&8.8&14.2&10.6\\
        I18~\cite{Ikehata2018}&Supervised&Calibrated&2.2&4.1&7.9&4.6&8.0&7.3&14.0&5.4&6.0&12.6&7.2\\
        CW20~\cite{Chen2020}&Supervised&Calibrated&2.7&7.7&\textbf{7.5}&4.8&6.7&7.8&12.4&6.2&7.2&10.9&7.4\\
        LB21~\cite{Logothetis2021}&Supervised&Calibrated&2.0&\textbf{3.5}&7.6&\textbf{4.3}&4.7&\textbf{6.7}&13.3&4.9&5.0&9.8&6.2\\
        LL22a~\cite{Li2022a}&Sup.+Self-Sup.&Calibrated&2.4&3.6&8.0&4.9&4.7&\textbf{6.7}&14.9&6.0&5.0&8.8&6.5\\
        \midrule
        CH19~\cite{Chen2019}&Supervised&Uncalibrated&2.8&6.9&9.0&8.1&8.5&11.9&17.4&8.1&7.5&14.9&9.5\\
        CW20~\cite{Chen2020}&Supervised&Uncalibrated&2.5&5.6&8.6&7.9&7.8&9.6&	16.2&7.2&7.1&14.9&8.7\\
        KK21~\cite{Kaya2021}&Sup.+Self-Sup.&Uncalibrated&3.8&6.0&13.1&7.9&10.9&11.9&25.5&8.8&10.2&18.2&11.6\\
        LL22b~\cite{Li2022b}&Sup.+Self-Sup.&Uncalibrated&\textbf{1.2}&3.8&9.3&4.7&5.5&7.1&14.6&6.7&6.5&10.5&7.0\\
        TR22~\cite{Tiwari2022} (K=2)&Self-Sup.&Uncalibrated&6.3&9.7&14.5&9.9&11.1&14.2&26.1&10.7&12.1&19.9&13.4\\
        \midrule
        I22~(UniPS)~\cite{Ikehata2022}&Supervised&Universal&4.9&9.1&19.4&13.0&11.6&24.2&25.2&10.8&9.9&18.8&14.7\\
        \rowcolor{gray!25}Ours&Supervised&Universal&1.5&3.6&\textbf{7.5}&5.4&\textbf{4.5}&8.5&\textbf{10.2}&\textbf{4.7}&\textbf{4.1}&\textbf{8.2}&\textbf{5.8}\\
        \midrule
        \midrule
        Ours (K=64)&Supervised&Universal&1.5&3.6&7.6&5.5&4.6&8.6&10.2&4.7&4.1&8.3&5.9\\
        Ours (K=32)&Supervised&Universal&1.5&3.6&7.7&5.5&4.7&8.6&10.4&4.8&4.2&8.4&5.9\\
        Ours (K=16)&Supervised&Universal&1.5&3.8&7.7&6.0&4.8&8.5&10.8&4.9&4.4&8.7&6.1\\
        Ours (K=8)&Supervised&Universal&1.6&4.0&8.2&6.3&5.2&8.4&11.5&5.2&4.8&9.4&6.5\\
        Ours (K=4)&Supervised&Universal&1.7&4.1&10.0&8.6&6.3&9.0&14.1&6.1&5.9&11.4&7.7\\
        Ours (K=2)&Supervised&Universal&1.9&6.8&14.4&13.6&8.3&12.8&21.2&9.0&9.2&16.9&11.4\\
    \bottomrule
    \label{table:diligent}
    \end{tabularx}
    \vspace{-20pt}
\end{table*}
\noindent\textbf{Training Details:}
Our network was trained on the PS-Mix dataset from scratch using the AdamW optimizer and a step decay learning rate schedule ($\times 0.8$ every ten epochs) with learning-rate warmup during the first epoch. A batch size of 8, an initial learning rate of 0.0001, and a weight decay of 0.05 were used. The number of input training images was randomly selected from $3$ to $6$ for each batch~\footnote{Six is the maximum number that can fit on our GPU.}. In our work, we chose ConvNeXt-T~\cite{Liu2022} as our backbone due to its simplicity and efficiency, which is better than recent ViT-based architectures~\cite{Dosovitskiy2020,Liu2021Swin,Liu2022Swin} with comparable performance. The training loss was the MSE loss, which computes the $\ell_2$ errors between the prediction and ground truth of surface normal vectors. Additional information, such as network architectures and feature dimensions, is provided in the appendix.
\\\\
\noindent\textbf{Evaluation and Time:} The accuracy is evaluated based on the mean angular errors (MAE) between the predicted and true surface normal maps, measured in degrees. Training is conducted on
four NVIDIA A100 cards for roughly three days. The inference time of our method depends on the number and resolution of input images. In the case of $16$ input images at a resolution of $512\times 512$, it takes a few seconds excluding I/O on a GPU. While the computational cost will vary almost linearly with the number of images, this is significantly more efficient than recent neural inverse rendering-based methods~\cite{Li2022a,Li2022b,Zhang2022,Munkberg2022}.
\subsection{Ablation Study}
Firstly, we perform ablation studies to evaluate the individual contributions of our scale-invariant spatial-light feature encoder and non-local interaction with pixel-sampling transformer across varying sample sizes. To quantitatively compare performance under various lighting conditions, we utilize the PS-Wild-Test dataset~\cite{Ikehata2022}, which contains 50 synthetic scenes rendered under three distinct lighting setups: directional, environmental, and a mixture of both. In~\Tref{table:ablation} and~\Fref{fig:ablation}, we compare our method with different configurations against~\cite{Ikehata2022}. Note that without the scale-invariant encoder and non-local interaction (\ie, the baseline), our method is nearly equivalent to~\cite{Ikehata2022}, except for some minor differences (\eg, backbone architecture, number of Transformer blocks in the encoder). We observe that the baseline method trained on our PX-Mix dataset improves performance for scenes under directional lighting, suggesting that one of the primary reasons why~\cite{Ikehata2022} was ineffective under directional lights was due to bias in the PS-Wild dataset. Accounting for non-local interaction of aggregated features enhances reconstruction accuracy, even with a small number of samples (\eg, $m$=32), as clearly illustrated in~\Fref{fig:ablation}-(top). Although accuracy improved as the number of samples increased, as expected, performance gains plateaued beyond a certain number (\ie, $m$=2048). The efficacy of the scale-invariant spatial-light feature encoder was also confirmed. In~\Fref{fig:ablation}-(bottom), we observed that a significant difference in input resolution to the backbone between training and testing led to substantial performance degradation, which further validates the advantage of our method that maintains a constant input tensor shape.
\subsection{Evaluation under Directional Lighting}
\begin{figure}[t]
	\begin{center}
		\includegraphics[width=85mm]{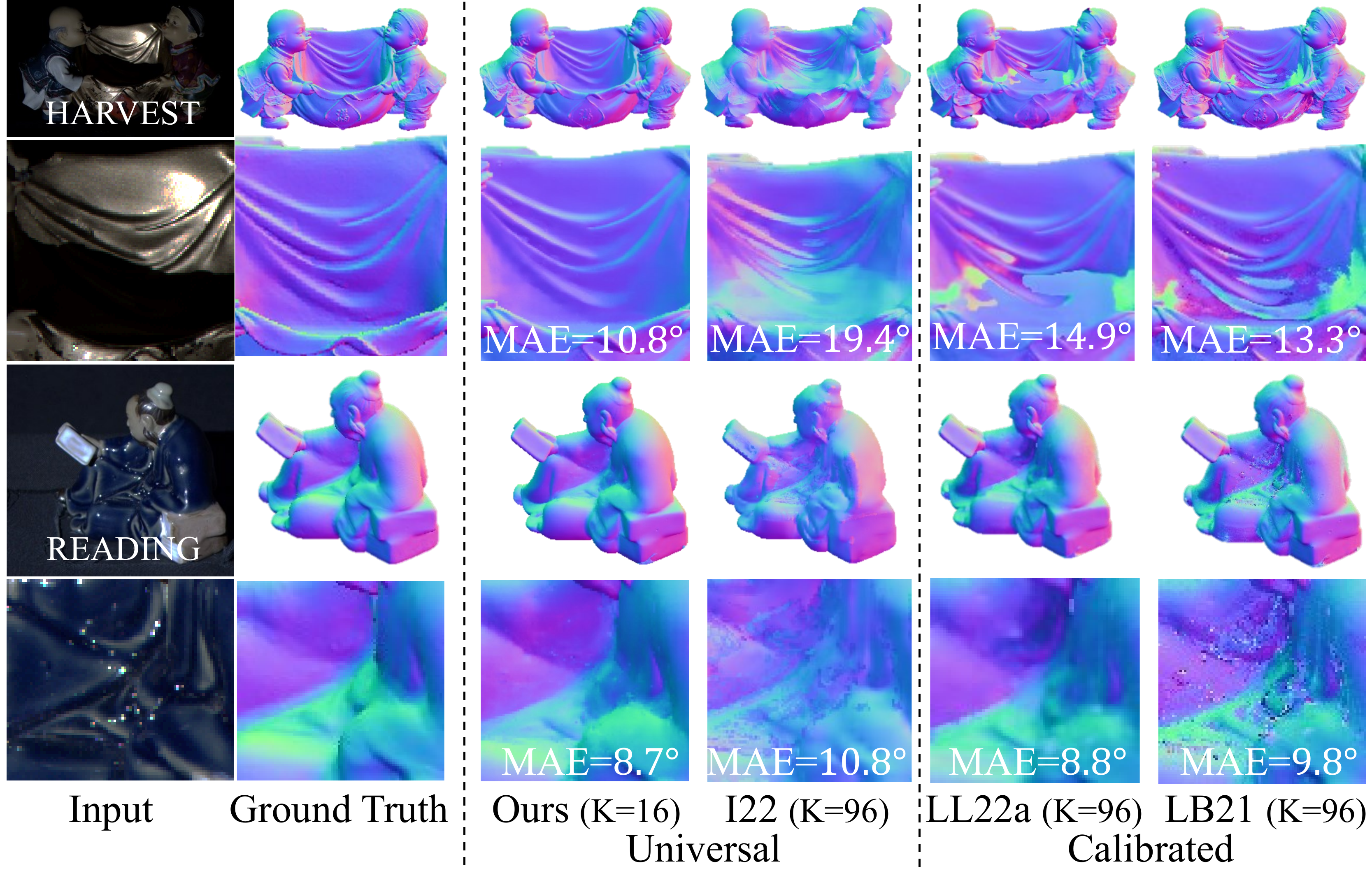}
	\end{center}
	\vspace{-15pt}
	\caption{Results for objects under a single directional lighting condition, including object masks.}
	\vspace{-10pt}
	\label{fig:diligent}
\end{figure}
\vspace{4.0pt}

\noindent\textbf{DiLiGenT Evaluation:} We first evaluate our method on the DiLiGenT benchmark~\cite{Shi2016}. Each dataset provides 96 612x512 16-bit HDR images, captured under known single directional lighting. The object mask and true surface normal map are available. In addition to UniPS~\cite{Ikehata2022}, we also compare our method with calibrated~\cite{Woodham1980,Ikehata2014a,Ikehata2014b,Ikehata2018,Chen2020,Logothetis2021,Li2022a} and uncalibrated~\cite{Chen2019,Chen2020,Kaya2021,Li2022b,Tiwari2022} photometric stereo algorithms specifically designed for single directional lighting. Calibrated methods include both pixelwise~\cite{Woodham1980,Ikehata2014a,Ikehata2014b,Ikehata2018,Logothetis2021} and image-wise~\cite{Chen2020,Li2022a} approaches. All uncalibrated methods are image-wise. We consider~\cite{Kaya2021,Li2022a,Li2022b} as a combination of supervised and unsupervised learning, as pretrained models were used as a starting point for lighting prediction. To evaluate the valid number of input images, we compare our method with different numbers of input images (results are averaged over 10 random trials).

The results are illustrated in~\Tref{table:diligent}. Impressively, our method, which does not assume a specific lighting model, outperforms state-of-the-art calibrated methods designed for directional lights (LB21~\cite{Logothetis2021}, LL22a\cite{Li2022a}). Furthermore, unlike conventional photometric stereo methods, the proposed method does not experience significant performance degradation even when the number of input images is reduced; it maintains state-of-the-art results even with only 8 images. The proposed method ($K=2$) also surpasses TR22~\cite{Tiwari2022}, which is specialized for two input images.

\begin{figure}[t]
	\begin{center}
		\includegraphics[width=85mm]{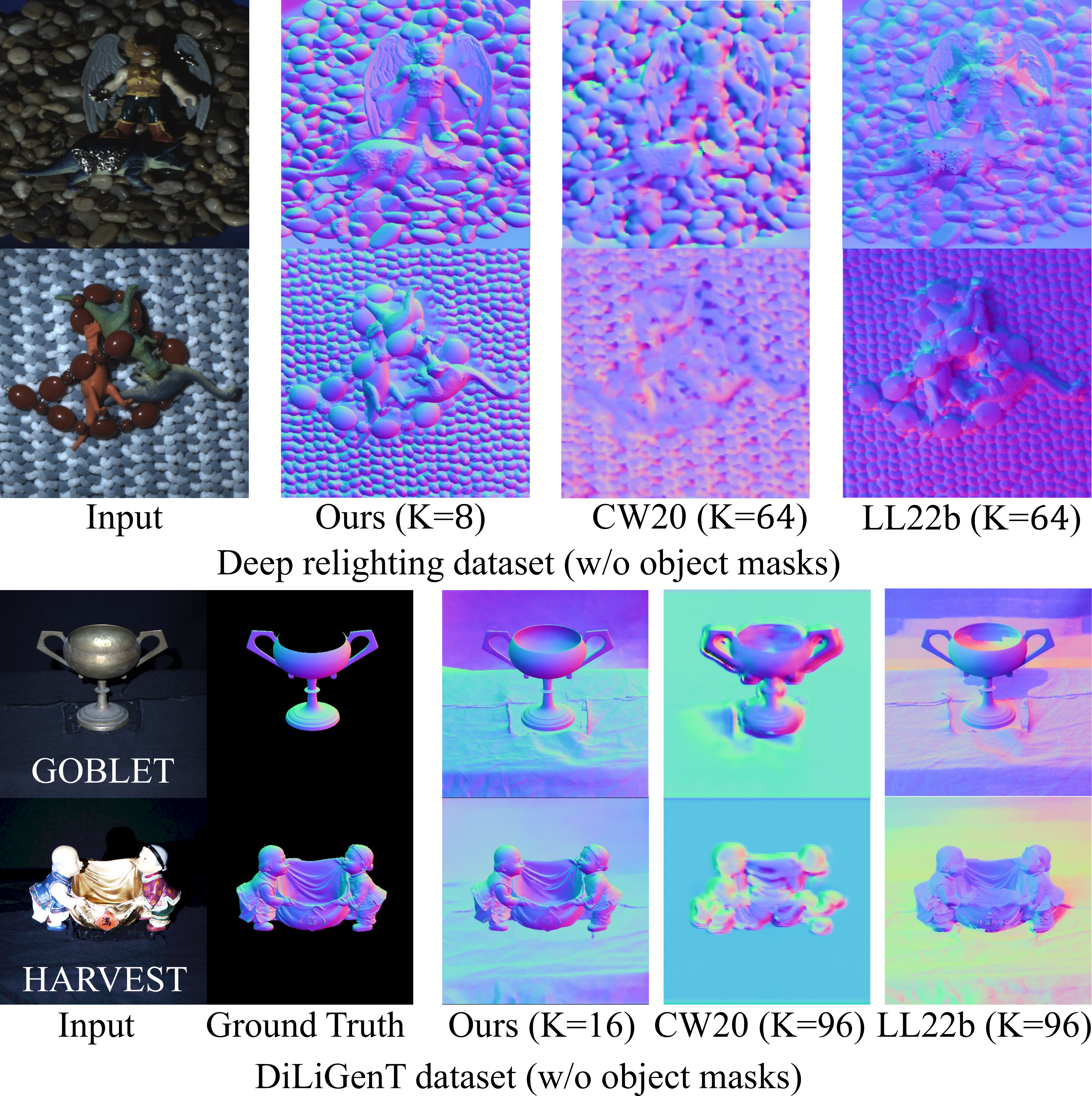}
	\end{center}
	\vspace{-20pt}
	\caption{Results for scenes under a single directional lighting condition, excluding object masks.}
	\vspace{-10pt}
	\label{fig:no_mask}
\end{figure}

Recovered normal maps of HARVEST and READING are shown in~\Fref{fig:diligent}. These objects are considered the most challenging in the benchmark due to their highly non-convex geometry. As expected, the state-of-the-art pixelwise calibrated method (LB21~\cite{Logothetis2021}) can recover finer surface details, while the state-of-the-art image-wise calibrated method (LL22a~\cite{Li2022a}) can recover more globally consistent results. However, both of them struggle to recover the non-convex parts of the objects accurately. On the other hand, our method can recover both surface details and overall shape without apparent difficulty, even with a much smaller number of images (\ie, K=16). As expected, the performance of I22~\cite{Ikehata2022} is severely lacking.
\vspace{4.0pt}\\
\noindent\textbf{Evaluation without Object Mask:} To demonstrate that our method does not require an object mask, we applied it to two real scenes from a deep relighting work~\cite{Xu2018}, each containing 530 8-bit integer images at a resolution of 512x512, captured under unknown single directional lighting using a gantry-based acquisition system. The object mask and true surface normal map are unavailable. We compared our method with state-of-the-art uncalibrated methods (CW20~\cite{Chen2020} and LL22b~\cite{Li2022b}) and displayed the results in~\Fref{fig:no_mask} (top). Unlike the uncalibrated methods that struggled to recover accurate lighting directions, our proposed method successfully captured object boundaries without masks, even in complex scenes with significant global illumination effects, and consistently recovered normals across the entire image. We further evaluated our method on DiLiGenT scenes without masks, as illustrated in~\Fref{fig:no_mask} (bottom). While existing methods that assume an object mask produced highly inaccurate surface normal maps, our proposed method recovered more plausible normals with fewer images (\ie, K=16 vs K=96).
\subsection{Evaluation under Spatially-varying Lighting}
\begin{figure}[t]
	\begin{center}
		\includegraphics[width=85mm]{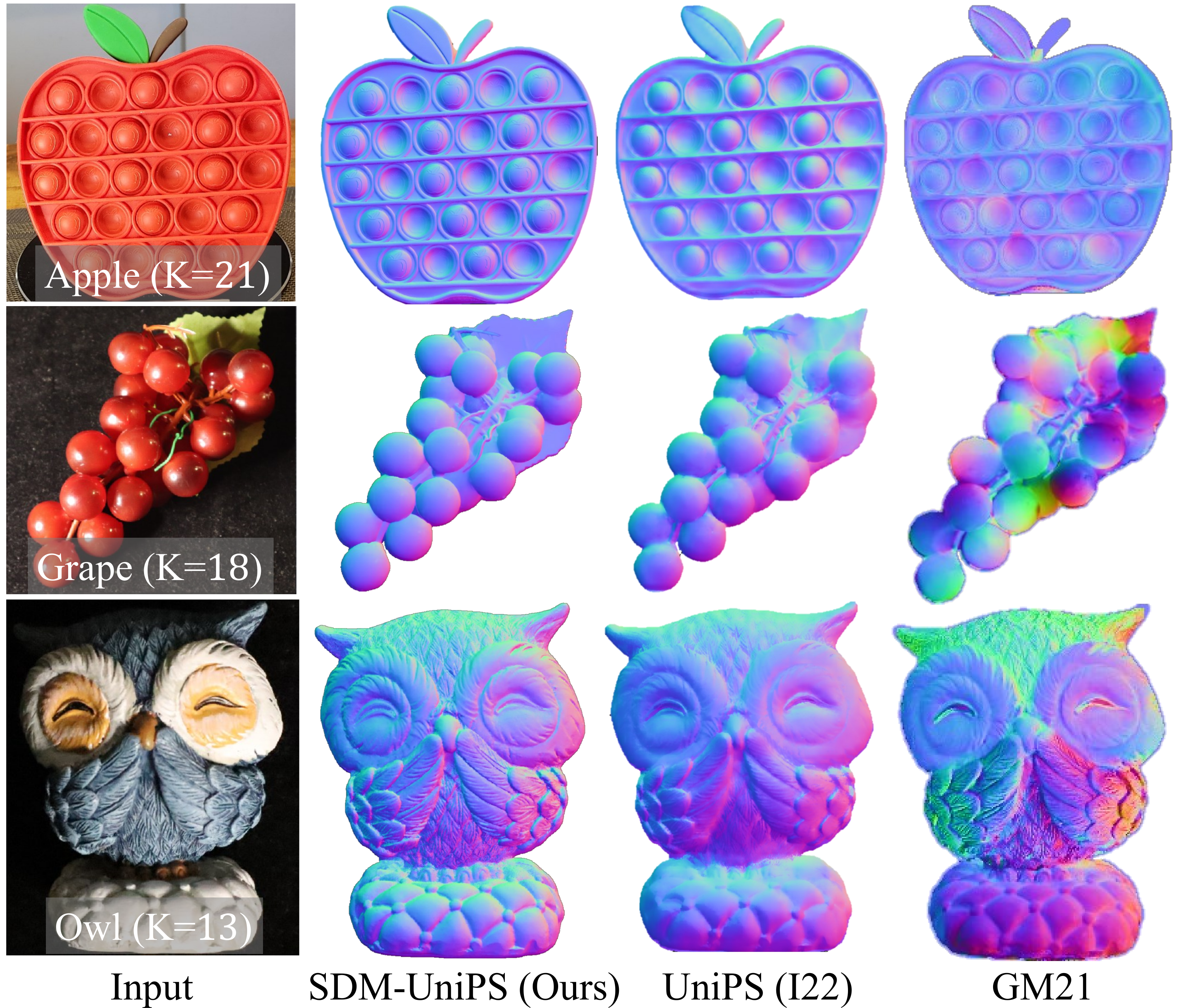}
	\end{center}
	\vspace{-20pt}
	\caption{Qualitative comparison on images under spatially-varying lighting conditions with object masks~\cite{Ikehata2022}.}
	\vspace{-10pt}
	\label{fig:cvpr2022}
\end{figure}
Our method is evaluated on challenging scenes with spatially-varying lighting conditions, comparing it to the first universal network (UniPS)~\cite{Ikehata2022} and a state-of-the-art uncalibrated photometric stereo method (GM21)~\cite{Guo2021} on a dataset provided by~\cite{Ikehata2022}. We test three objects (Apple, Grape, and Owl). While GM21~\cite{Guo2021} fails and I22~\cite{Ikehata2022} loses details, our method, using a scale-invariant spatial-light feature encoder and non-local interaction, produces accurate results.

\begin{figure}[t]
	\begin{center}
		\includegraphics[width=85mm]{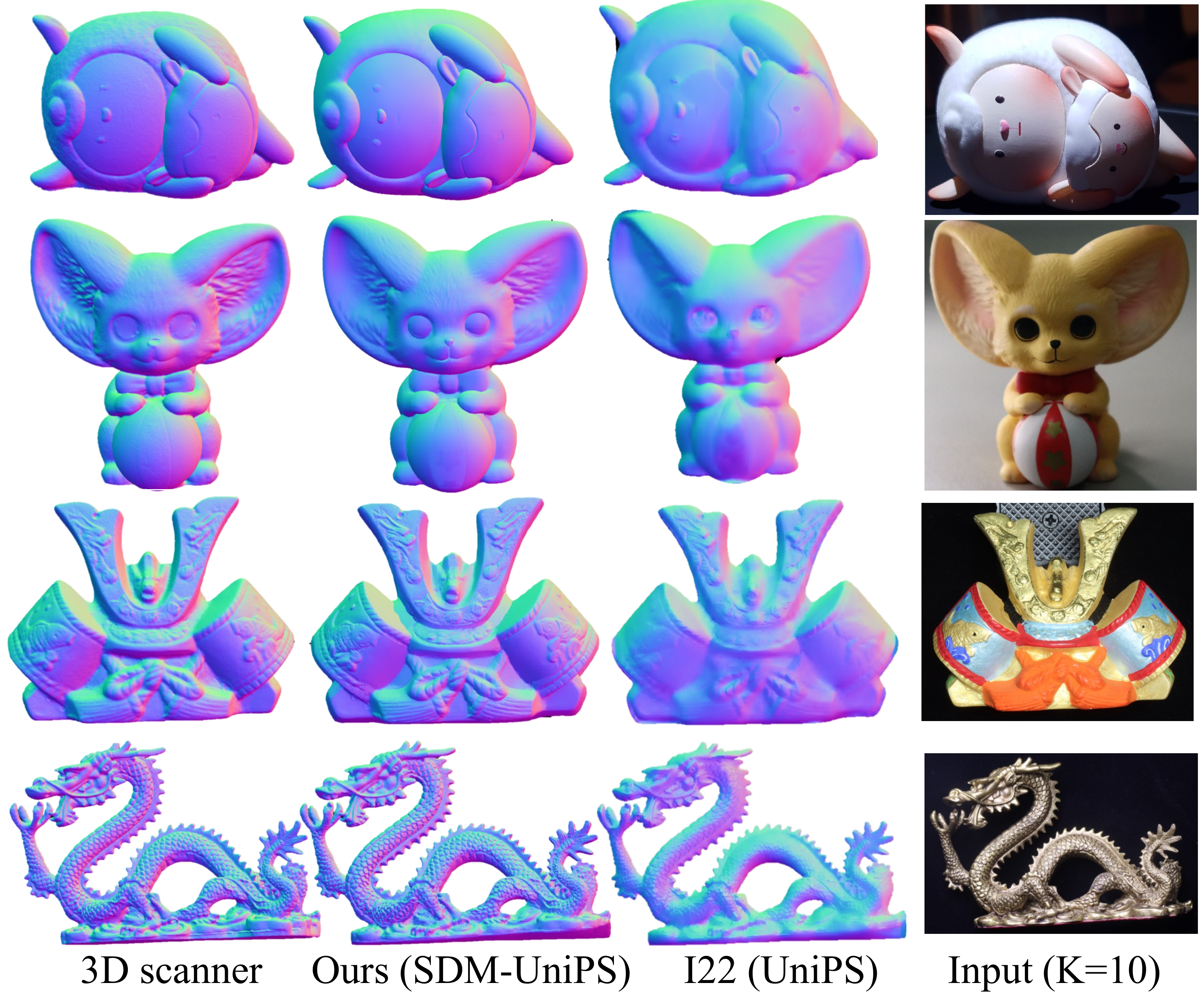}
	\end{center}
	\vspace{-20pt}
	\caption{Qualitative comparison with 3-D scans.}
	\vspace{-10pt}
	\label{fig:quantitative_real}
\end{figure}

In~\Fref{fig:quantitative_real}, we subjectively compare our method using four objects with normal maps obtained from a 3D scanner. We align the scanned normal map to the image using MeshLab's mutual information registration filter~\cite{Meshlab}, as in~\cite{Shi2016}. Our method recovers higher-definition surface normal maps than the 3D scanner (EinScan-SE) and performs well regardless of surface material. Photometric stereo performance improves with increased digital camera resolution, suggesting that 3D scanners may struggle to keep up.

\begin{figure}[t]
	\begin{center}
		\includegraphics[width=85mm]{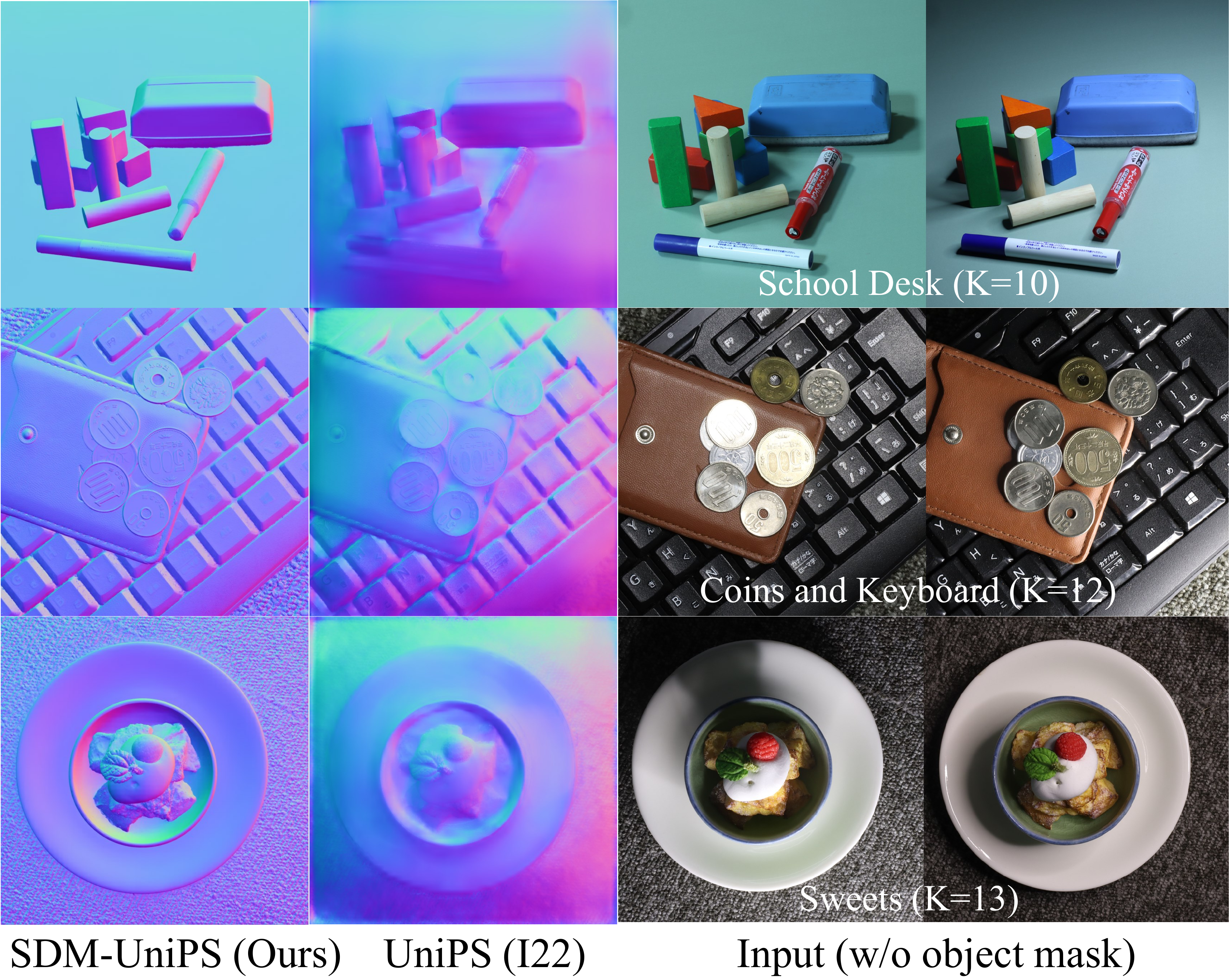}
	\end{center}
	\vspace{-20pt}
	\caption{Surface normal recovery from images under spatially-varying lighting conditions without object masks.}
	\vspace{-10pt}
	\label{fig:no_mask_sv}
\end{figure}

Lastly, we demonstrate surface normal prediction for complex non-convex scenes without masks under challenging lighting conditions in Figure~\ref{fig:no_mask_sv}. We apply our method to three extremely challenging datasets: School Desk, Coins and Keyboard, and Sweets. School Desk is a complex scene with simple objects, non-uniform lighting, and cast shadows, making surface normal map recovery difficult. Coins and Keyboard features multiple planar objects of various materials. Sweets is a challenging scene with abundant inter-reflections and cast shadows. As demonstrated, the proposed method successfully recovers uniform surface normals, largely unaffected by shadows, and effectively reconstructs the surface micro-shape, demonstrating its scalability and detail preservation.
\section{Conclusion}
In this paper, we presented a scalable, detailed, and mask-free universal photometric stereo method. We demonstrated that the proposed method outperforms most calibrated and uncalibrated methods in the DiLiGenT benchmark. In addition, the comparison with the only existing method~\cite{Ikehata2022} for the universal task showed a significant improvement over it.

However, several challenges still remain. Firstly, although we have observed that the proposed method works robustly for versatile lighting conditions, we found that our method is not very effective when the lighting variations are minimal. Secondly, the proposed method can easily be extended beyond normal map recovery by replacing the loss and data. In reality, we have attempted to output BRDF parameters for materials. However, due to fundamental ambiguities, it is difficult to evaluate the recovered BRDF parameters. Please see the appendix for further discussions of these limitations and a variety of additional results to better understand this study.
{\small
\bibliographystyle{ieee_fullname}
\bibliography{egbib}
}

\appendix
\newpage
\section*{Appendix A. Failure Cases and Better Imaging}
While the main text discusses only the theoretical aspects of the problem, this section discusses the practical aspects of our method by discussing in more detail the limitations regarding the input acquisition and how to capture better images for our method.
\vspace{5pt}

\noindent\textbf{Basics about image acquisition.} Our image acquisition process is simple. Prepare a scene and take photos of it under different lighting conditions without moving a camera. The light source can theoretically be either active (\eg, using a hand-held light) or passive (\eg, mounting a camera and an object on the same board and moving them around) as long as sufficient changes in illumination occur. Realistically, the most probable situation may involve a combination of dynamic active lights in a static environment.

Our method has no restrictions on the size of scenes. On the other hand, since the proposed method assumes an orthographic camera, extreme projection distortion is not considered. However, as a common practice, the view directions of a perspective projection camera become more parallel with each other around the central field of view, so using only the central region of a sufficiently high-resolution image is not problematic for practical purposes. Throughout the papers (\ie, main and supplementary), we used either a 45mm or 200mm focal length camera based on the object size to capture 4000x4000 images, of which the central 2048x2048 area was used in the preprocessing as described in the main paper.
\vspace{5pt}

\noindent\textbf{Failure cases and possible solutions.}
\begin{figure}[t]
	\begin{center}
		\includegraphics[width=80mm]{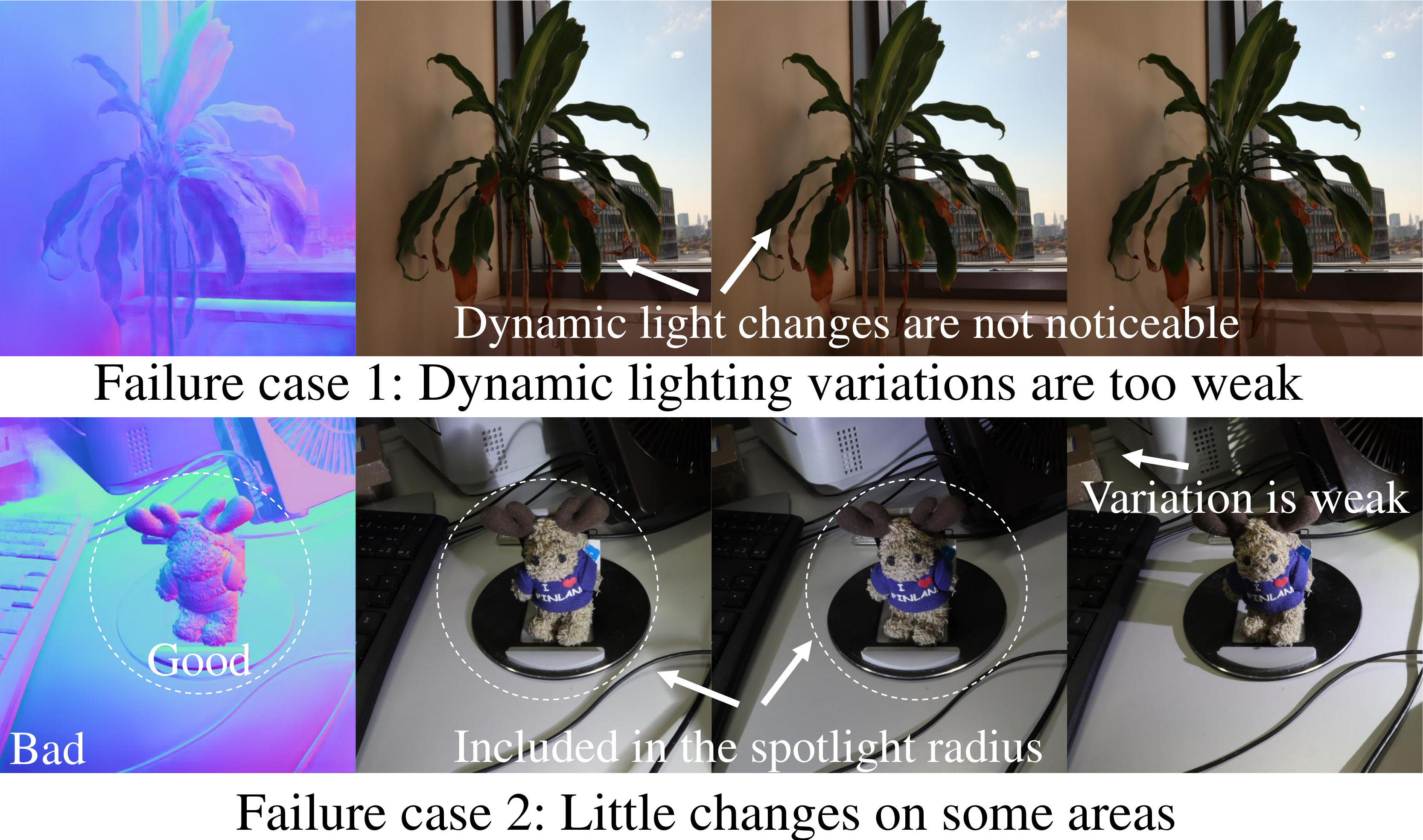}
	\end{center}
	\vspace{-10pt}
	\caption{Failure cases. The performance of the proposed method degrades significantly when changes in the illumination environment cannot be observed, whether in part or in the entire image.}
	\vspace{-10pt}
	\label{fig:failure}
\end{figure}
We observed two major cases of failure in the course of our experiments as illustrated in~\Fref{fig:failure}. First, the performance drastically degrades if the unmasked region contains areas where no or little illumination change exists because our training data (PS-Mix) contains no cases where the light source condition doesn't change or is very weak from image to image in any regions of the image. For example, when a spotlight light is illuminated on an object, the surface normal recovery could fail if the image contains many areas that are not included in the light diameter. Another common case is that the intensity of the dynamic light source is very weak compared to the static one therefore illumination changes between images are scarce. This tends to occur when the method is applied during the daytime or when trying to recover large scenes of wide depth range. 

There are various possible ways to improve this, such as improving the training data by including such cases~(\eg, spot light rendering) or adding a mechanism to identify and ignore regions where light source changes do not occur, but further discussion would go beyond what is allowed in the supplementary, so we leave these issues for the future work.
\vspace{5pt}

\noindent\textbf{The better choice of light source.} Based on the discussion above, a point light source or surface light source that can illuminate a wide area simultaneously may seem more appropriate, rather than a spotlight that tends to produce areas that are not clearly illuminated. Generally speaking, when automatic exposure control is turned on, the tonal resolution is degraded to increase dynamic range when very dark and bright areas are mixed together. On the other hand, when the entire image is bright, it is possible to represent enough information within a narrow dynamic range, resulting in less image noise. Therefore, to improve the quality of a captured image, it is essential to make the irradiance uniform across the image.

\begin{figure}[t]
	\begin{center}
		\includegraphics[width=80mm]{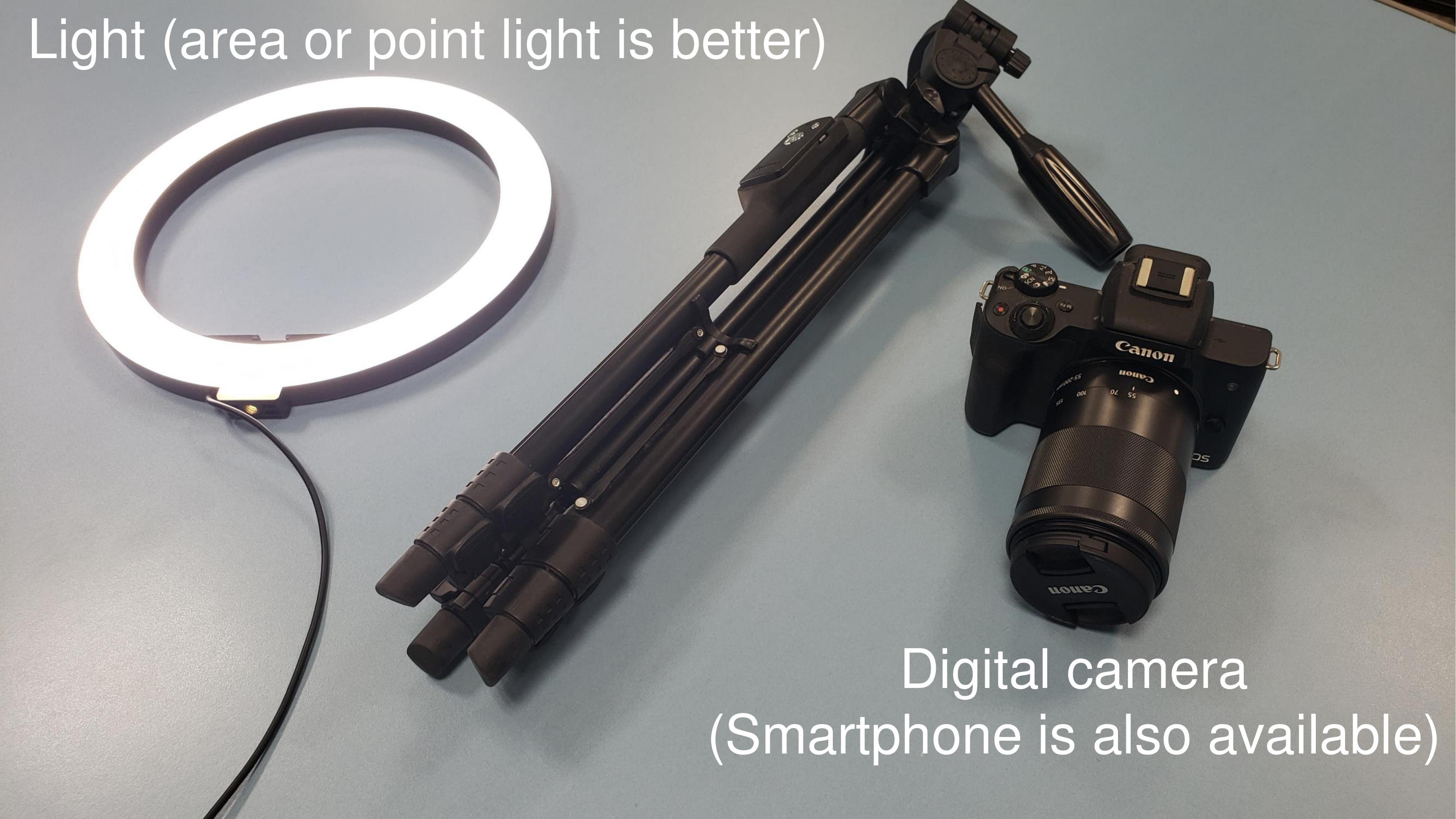}
	\end{center}
	\vspace{-10pt}
	\caption{Our acquisition setup simply needs a movable light source and a camera.}
	\vspace{-10pt}
	\label{fig:tools}
\end{figure}

Empirically, we have found that using a ring light for selfies or a smartphone/tablet screen as a light source are the two most effective methods available to us that meet the above conditions. Of these, the selfie-light, which provides sufficient light and is easy to handle, was used in many of the experiments in this paper. The tools used in this work are shown in~\Fref{fig:tools}. Since no calibration of the light source or camera is necessary, all that is required are a single light source, a single camera, and target objects.
\vspace{5pt}

\noindent\textbf{When a mask is necessary and when it is not.} Basically, the proposed method does not require a mask. As one may have noticed from the paper's results, our method is capable of preserving the depth discontinuities of objects without a mask to a level that is not possible with any existing methods. There are three main factors that make this possible. First, unlike the existing dataset (\ie, PS-Wild), our PS-Mix consists of multiple overlapping objects, and learning is performed without explicitly providing their boundaries. Second, our method is more robust than all existing methods against inter-reflections and cast shadows that occur at depth boundaries. Third, during global interactions of the aggregated features, no local interaction is performed unlike existing methods, so no over-smoothing occurs.

However, there are some cases where the object mask is helpful. The first case is simply when one wishes to recover only the shape of a particular object in a scene. The second case is when we want to explicitly "hide" areas with little lighting variation. During network training, a ground truth mask is always given simultaneously, and the loss function is computed only from pixels in the mask. Consequently, information outside the mask is not taken into account in the prediction during training. In other words, a mask can be used to intentionally hide areas from the network where lighting variations are weak. For this purpose, the mask does not need to follow the contour of the object; a bounding box-like specification is sufficient.
\vspace{5pt}\\
\noindent\textbf{Other points to note on photography.} We found that there are other exceptional cases where our method does not work well. If the image correction is too strong, it will fail. For example, recent smartphones apply various image filters to improve the appearance of images after they are taken. As a result, the physically correct shading changes are destroyed. Also, while the proposed method basically does not require HDR (high-dynamic-range) images, it is not as robust with respect to too much over- and under- exposure. Fortunately, the automatic exposure control provided in recent digital cameras and smartphones is very effective to avoid the situations. Similarly to other photometric stereo methods, our method is also helpless with respect to an image blur and an accidental misalignment of images. The above problems can be easily solved by carefully tuning the camera, so they are not critical in practice.
\subsection*{Summary}In conclusion to this section, the following points should be kept in mind when taking photographs.
\begin{itemize}
\item To assume an orthographic camera, the object should be placed in the central field of view of a camera with a sufficiently large focal length.
\item Ensure that the illumination changes throughout the image. For this purpose, light sources that can illuminate a wide area, such as a point light or a surface light, are better than a spotlight. Alternatively, masks can be used to hide areas of weak illumination variation.
\item Turn off software image correction, increase the depth of field to prevent blur, and ensure that the camera does not move while taking photographs.
\item The number of images can be small. If you need more, just add more. It only takes a few seconds.
\end{itemize}
\section*{Appendix B. Network Architecture Details}
Our entire framework consists of six sub-networks. The scale-invariant spatial-light encoder includes (a) a backbone network for the imagewise feature extraction, (b) a Transformer network for the pixelwise interaction along the light-axis and (c) a feature pyramid network for the fusion of hierarchical feature maps. And in the pixel-sampling Transformer, there are (d) a Transformer network for the feature aggregation along the light-axis and (e) a Transformer network for the feature interaction along the spatial-axis. Finally, we have (f) a MLP for the surface normal prediction. In this section, we detail each network architecture.
\vspace{5pt}\\
\noindent\textbf{Backbone}: In our scale-invariant spatial-light encoder, each sub-tensor (\ie, concatenation of a sub-image and a sub-mask) is independently input to ConvNeXt~\cite{convnext} which is a modernized ResNet~\cite{He2016} like architecture taking inspiration from the recent Vision Transformer~\cite{Dosovitskiy2020,Liu2021Swin}. The variants of ConvNeXt differ in the number of channels C, and the number of ConvNeXt blocks B in each stage. We here chose the following configuration.
\begin{itemize}
    \item  ConvNeXt-T: C = (96, 192, 384, 768), B = (3, 3, 9, 3)
\end{itemize}
The ConvNeXt block includes 7x7 depthwise convolution, 1x1 convolution with the inverted bottleneck design (4x hidden dimension) and 1x1 convolution to undo the hidden dimension. Between convolutions, layer normalization~\cite{Xiong2020} and GeLU~\cite{Hendrycks2016} activation are placed. The output of ConvNeXt is a stack of feature maps of (B x 96 x R/4 x R/4), (B x 192 x R/8 x R/8), (B x 384 x R/16 x R/16) and(B x 768 x R/32 x R/32) where B is the batch size and R is the input sub-tensor size as defined in the main paper. 
\vspace{5pt}\\
\noindent\textbf{Transformer (interaction along light-axis)}:
Given hierarchical feature maps from the backbone network, we pixelwisely apply Transformer~\cite{Vaswani2017} to features of individual scales along the light-axis as with~\cite{Ikehata2022}. We chose the number of channels in a hidden layer C, and the nuber of Transformer blocks B as follow.
\begin{itemize}
    \item  Transformer: C = (96, 192, 384, 768), B = (0, 1, 2, 4)
\end{itemize}
The Transformer block projects the input feature to query, key and value vectors whose dimensions are same with the input ones. They are then passed to a multi-head self-attention (the number of heads is $8$) with a soft-max and a feed-forward network with two linear layers whose dimensionality of input and output layers is same but one of the inner layer is twice of the input. A residual connection around each of the two sub-layers, followed by layer normalization~\cite{Xiong2020} and dropout ($p=0.1$).  
\vspace{5pt}\\
\noindent\textbf{Feature pyramid network}:
After the hierarchical feature maps pixelwisely interact with each other using Transformers, feature maps of different scales corresponding to each input image are fused with the feature pyramid network (\ie UPerNet~\cite{Xiao2018}) which was originally proposed for the semantic segmentation task. We simply used an implementation on MMSegmentation~\cite{MMSegmentation} without any modifications. The output feature size is (B x R/4 x R/4 x 256).
\vspace{5pt}\\
\noindent\textbf{Transformer (aggregation along light-axis)}:
Given $m$ pixel locations at the input coordinate system, we concatenate each pair of a raw observation and a bilinearly interpolated feature vector from the output of the feature pyramid network to a vector whose dimension is 259 (\ie, 256+3). The feature aggregation network takes $K$ sets of 259-dim feature vectors at the same location as input and perform two Transformer blocks of C=256 (shrunk from 259 to 256 by QKV projection). The output feature is further concatenated with the raw observation and each 259-dim feature vectors are again fed to another three Transformer blocks of C=256. Then, the output K feature vectors are passed to PMA~\cite{Lee2019} where the number of elements in a set was shrunk from $K$ to one using another Transformer block of C=384.
\vspace{5pt}\\
\noindent\textbf{Transformer (interaction along spatial-axis)}: At the final step of the pixel-sampling Transformer module, we perform two Transformer blocks (C=384) to communicate features among the $m$ locations. The n\"aive self-attention requires O($m^2$) memory consumption, however $m$ (\ie, number of pixel samples) is much larger than $K$ (\ie, number of input images), which makes increasing sample size difficult. Therefore, we instead used the O($m$) implementation of the self-attention by~\cite{Rabe2021} to tackle this problem (Note that the computational cost doesn't change). 
\vspace{5pt}\\
\noindent\textbf{Normal prediction network}: The surface normal predictor is a MLP with one hidden layer whose feature dimension shrank as $384\rightarrow 192 \rightarrow 3$ and the norm of the output vector is normalized to be a unit surface normal vector at the location.
\begin{figure*}[t]
	\begin{center}
		\includegraphics[width=160mm]{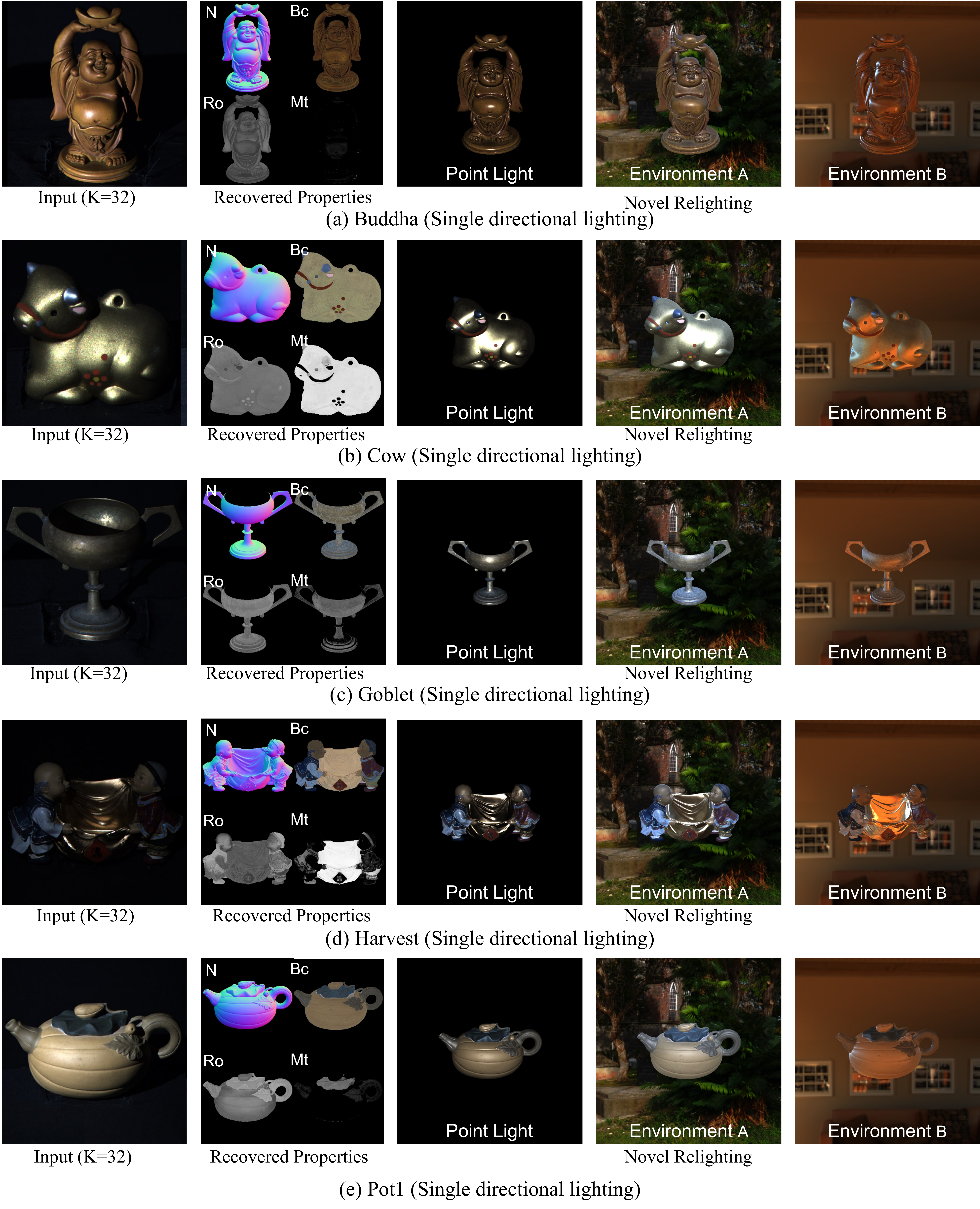}
	\end{center}
	\vspace{-10pt}
	\caption{Reflectance recovery and novel relighting of scenes under directional lightings.}
	\vspace{-10pt}
	\label{fig:relighting1}
\end{figure*}
\begin{figure*}[t]
	\begin{center}
		\includegraphics[width=160mm]{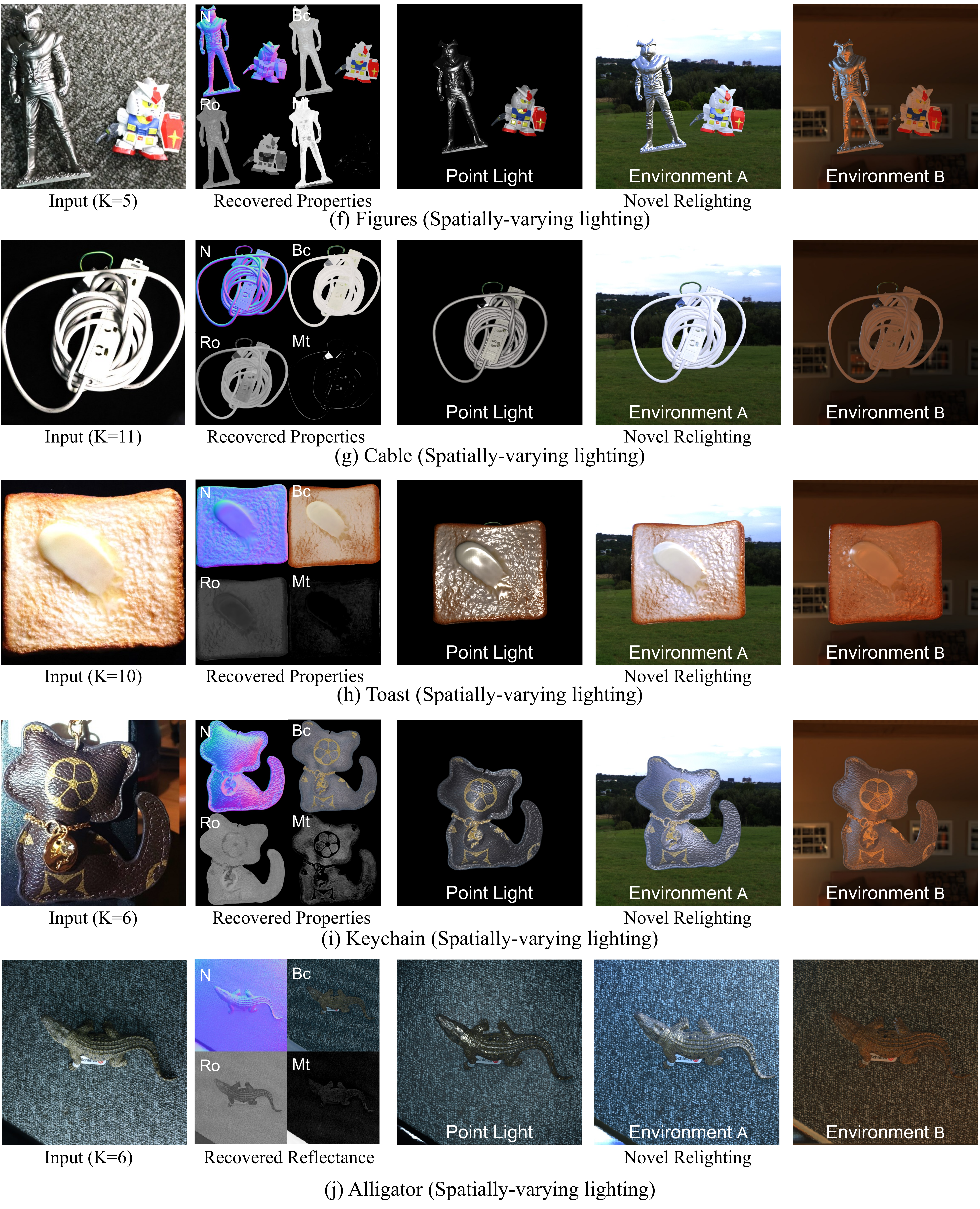}
	\end{center}
	\vspace{-10pt}
	\caption{Reflectance recovery and novel relighting of scenes under spatially-varying illuminations.}
	\vspace{-10pt}
	\label{fig:relighting2}
\end{figure*}
\begin{figure}[t]
	\begin{center}
		\includegraphics[width=80mm]{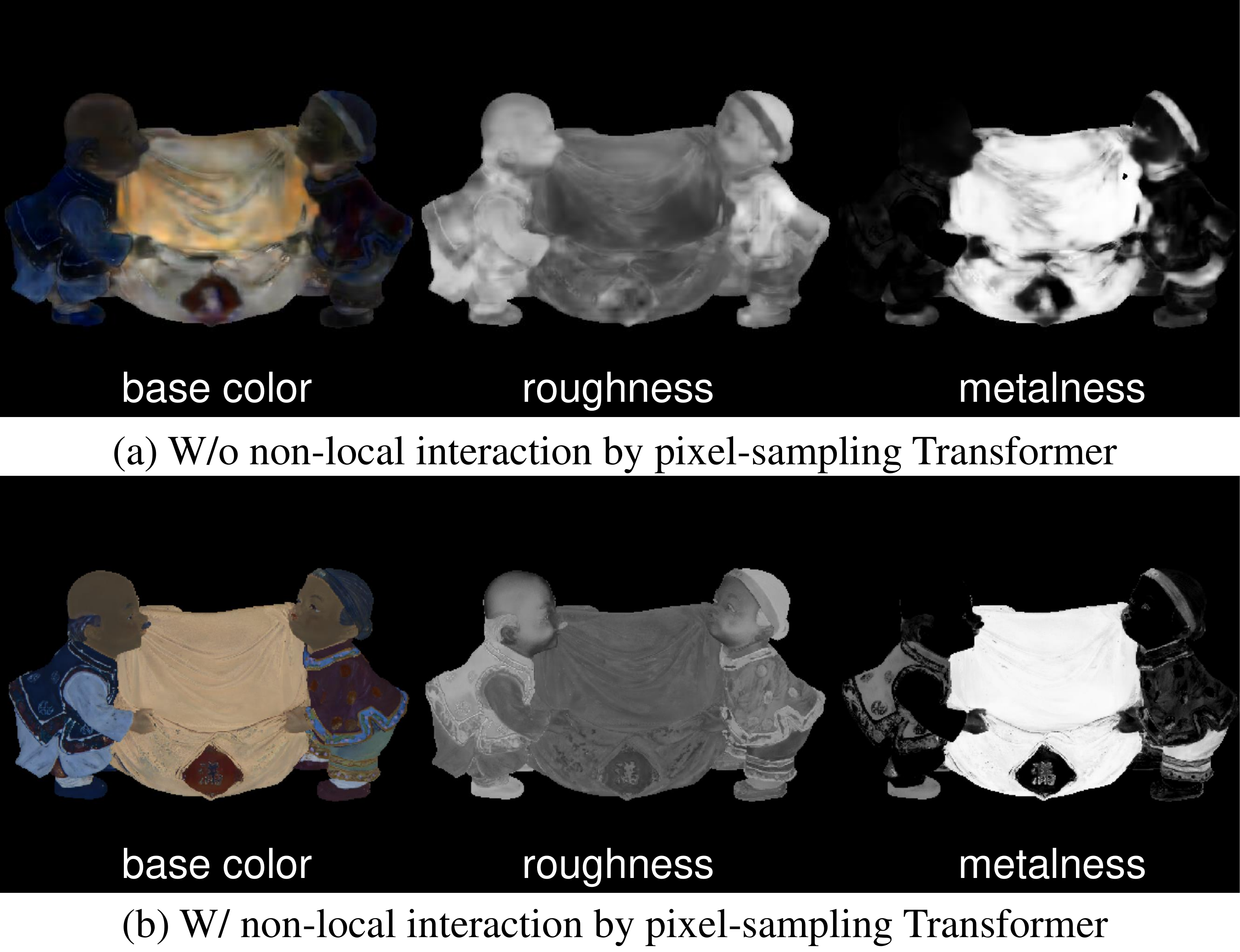}
	\end{center}
	\vspace{-10pt}
	\caption{We compared the results of reflectance recovery w/ and w/o our pixel sampling Transformer. As we observe, the non-local interaction among aggregated features seem to be critical in the surface reflectance recovery.}
	\vspace{-10pt}
	\label{fig:relighting3}
\end{figure}
\section*{Appendix C. BRDF Parameter Recovery}
As highlighted in the main paper's conclusion, the proposed framework extends beyond surface normal recovery and can be readily applied to surface reflectance recovery by merely substituting the training data and loss function. This versatility enables the proposed method to render the target scene under novel lighting environments, or in other words, achieve novel relighting. However, recovering surface reflectance from images in an uncalibrated setup poses a fundamental ambiguity due to the countless possible combinations of illumination and reflectance, such as a red surface under white light or a white surface under red light. This complexity makes objective evaluation nearly unattainable.

Given the challenges in evaluation, we opted not to feature the results of reflectance recovery in the main paper. Instead, we present them here to demonstrate that our method is not confined to surface normal recovery. Our PS-Mix dataset already incorporates base color, roughness, and metalness maps from AdobeStock~\cite{AdobeStock}, which were employed to render images. We simply utilize these maps as training data and train our network using Mean Squared Error (MSE) losses between the predicted and provided base color, roughness, and metalness maps. Once the surface reflectance parameters and surface normals are recovered, we can render images of the scene under novel lighting conditions using any physically-based renderer, such as Blender~\cite{Blender}.
\vspace{5pt}\\
\noindent\textbf{Implementation details}: We implemented two separate networks for surface normal map recovery and base color, roughness, and metalness maps recovery, respectively. We observed that training a single network for both tasks slightly degraded performance. The network architecture and training methodology were identical to those described in the main paper, with the only difference being the training data and loss functions.
\\
\noindent\textbf{Reflectance representation}: The images in both PS-Wild~\cite{Ikehata2022} and our PS-Mix were rendered using the dichromatic Bidirectional Reflectance Distribution Function (BRDF)\cite{DisneyPrincipledBSDF}, which is commonly assumed in physically-based rendering of materials. This BRDF is a combination of the diffuse, specular, and metallic BRDFs, controlled by three parameters: {\it base color} $\in\mathbb{R}^3$, {\it roughness} $\in\mathbb{R}$, and {\it metalness} $\in\mathbb{R}$ (all parameters are within the range $0$ to $1$). The diffuse BRDF includes a Schlick Fresnel factor and a term for diffuse retro-reflection whose color is determined by the {\it base color} parameter. The specular BRDF is the Cook-Torrance microfacet BRDF that uses the isotropic GGX (also known as Generalized-Trowbridge-Reitz-2) with a Smith masking-shadowing function. The {\it roughness} parameter controls the shape of the lobe, with smaller {\it roughness} values producing steeper specular lobes, i.e., more prominent specular highlights. The metallic BRDF uses the same specular BRDF, but the reflected light is colored with the {\it base color} parameter. The {\it metalness} parameter balances the weight between the dielectric (diffuse+specular) and metallic BRDFs. As per this definition, the metalness of a surface is primarily determined by the color of its specular reflection. In other words, if the predicted surface base color and the color of the specular reflection are similar, the surface is classified as metallic and assigned a corresponding metalness value. This can result in some black surfaces being classified as metallic, but it does not pose an issue in novel relighting since black surfaces are always represented by the same BRDF, regardless of their metalness value. For further details, please refer to\cite{DisneyPrincipledBSDF}.
\vspace{5pt}\\
\noindent\textbf{Results under directional lighting}:
In~\Fref{fig:relighting1}, we present the results of reflectance recovery from random 32 images of five objects in DiLiGenT~\cite{Shi2016}. For each object, we show one of the input images and the recovered surface normal (N), {\it base color} (Bc), {\it roughness} (Ro), and {\it metalness} (Mt) maps. We observe that the proposed method could cluster identical materials, even though we did not impose any physically-based constraints on reflectance properties based on prior knowledge, such as smoothness or sparsity of basis materials, which has been done in most existing works~\cite{Zhang2022,Zhang2022IRON,Li2022a,Li2022b}. We also observe successful separation between the surface color and shading effects. Furthermore, several objects in DiLiGenT have metallic paintings (\eg, Harvest and Cow), and our method correctly recovered the metalness values for these areas. To the best of our knowledge, our method is the first to recover physically plausible metalness parameters of non-convex scenes under unknown lighting conditions. 

Using the recovered BRDF parameters, we rendered the scenes under three different lighting conditions using the physically-based renderer~\cite{Blender}: a point light collocated with the camera position, outdoor environment lighting, and indoor environment lighting. While the unavoidable ambiguity of the problem setup makes quantitative evaluation impossible, we obtained highly plausible rendering results for each lighting condition. Note that all results were based on the surface normal map, not the surface meshes, so we cannot render global lighting effects such as cast shadows and inter-reflections. Our analysis of the results revealed an interesting observation: non-local interactions are more critical in recovering surface reflectance than surface normal. As shown in~\Fref{fig:relighting3}, we found that the recovery of material properties required a broad range of observations, including from low-frequency (diffuse) to high-frequency (specular) components. Reliable low-frequency information is almost sufficient for surface normal prediction, but it is not enough for recovering material properties, and focusing on a specific pixel is not adequate. The non-local interaction of aggregated features proved helpful in seeing different surface points of the same material for the recovery of surface reflectance, resulting in our outstanding results.
\vspace{5pt}\\
\noindent\textbf{Results under spatially-varying lighting}: In~\Fref{fig:relighting2}, we present the results of reflectance recovery and novel relighting from images under spatially-varying illuminations. The results demonstrate that the proposed method achieves physically plausible performance, overcoming the challenging conditions of each scene.

For instance, in the {\it Figures} dataset, a metallic-painted, non-planar object and a non-metallic, planar object exist in the same scene, but the network successfully reconstructed the normal map without distinguishing between these objects. Additionally, the {\it metalness} parameters were successfully recovered in the metallic-painted area, as demonstrated in the relighting results. The {\it Cable} objects have complex tangles of long, thin cables, and such geometries tend to produce ambiguous depth discontinuities when handled by existing methods. The proposed method not only accurately reconstructed these geometries but also recovered uniform {\it base color} without being affected by cast shadows or inter-reflections caused by non-convex geometries. For other objects such as {\it Toast}, {\it Keychain}, and {\it Alligator}, the proposed method successfully recovered a detailed surface normal map that preserved depth discontinuities accurately and produced perceptually plausible surface reflectance maps that were unaffected by shading and global illumination effects. The novel relighting results demonstrate that our results are of practical quality for the capture of surface attributes.

These results suggest that the proposed method is highly effective not only in surface normal recovery but also in surface reflectance recovery and novel relighting of the scene. However, we emphasize once again that we are aware that the results of this experiment are not objective, and further quantitative evaluation of surface reflectance recovery is left for future work.

\end{document}